\pdfoutput=1

\documentclass[11pt]{article}

\usepackage[]{ACL2023}
\usepackage{amsmath}
\usepackage{times}
\usepackage{latexsym}
\usepackage{newtxtext} 
\usepackage{newtxmath} 
\usepackage[T1]{fontenc}

\usepackage[utf8]{inputenc}
\usepackage{booktabs}
\usepackage{anyfontsize}
\usepackage{microtype}
\usepackage{tcolorbox}
\usepackage{inconsolata}
\usepackage{soul}
\usepackage{multirow}
\usepackage{epstopdf}
\usepackage{hyperref}
\usepackage{listings}
\usepackage{fancybox}
\usepackage{graphicx}
\usepackage{subcaption}
\usepackage{paralist}
\usepackage{ragged2e}
\usepackage{enumitem}
\usepackage{tcolorbox}
\makeatletter
\usepackage{xcolor}
\usepackage{algpseudocode}

\usepackage{color}
\usepackage{xcolor,colortbl}
\usepackage{balance}
\usepackage{threeparttable}

\usepackage[framemethod=TikZ]{mdframed}
\usepackage{tcolorbox}
\makeatletter
\newcommand{\mybox}[1]{%
  \setbox0=\hbox{#1}%
  \setlength{\@tempdima}{\dimexpr\wd0+13pt}%
  \begin{tcolorbox}[boxrule=0.5pt, colback=white, arc=4pt,
      left=6pt,right=6pt,top=6pt,bottom=6pt,boxsep=0pt]
    #1
  \end{tcolorbox}
}
\usepackage{tikz}
\usepackage{pgfplots}
\usetikzlibrary{pgfplots.statistics,calc}
\usepackage{color}
\usepackage{array}
\usepackage{amsmath}
\usepackage{centernot}
\usepackage{xspace}
\usepackage{url}
\usepackage{verbatim}
\usepackage{wrapfig}
\usepackage{tabularx}
\usepackage{bbding}
\usepackage{pifont}
\usepackage{wasysym}
\usepackage{cleveref}
\usepackage{makecell}

\makeatletter  
\newif\if@restonecol  
\makeatother  
\usepackage[linesnumbered,ruled,vlined]{algorithm2e}
\usepackage{algpseudocode}  
\newcommand{\tool}{\textit{RAG+CoE}}
\title{What External Knowledge is Preferred by LLMs? Characterizing and Exploring  Chain of Evidence in Imperfect Context for Multi-Hop QA}

\author{
\fontsize{10pt}{6pt}\selectfont
  Zhiyuan Chang\textsuperscript{\normalfont 
 1,2,3}   \hspace{0.5cm}
  Mingyang Li\textsuperscript{\normalfont 
 1,2,3}  \hspace{0.5cm}
  Xiaojun Jia\textsuperscript{\normalfont 4} \hspace{0.5cm}
  Junjie Wang\textsuperscript{\normalfont  1,2,3}
  \\
  \fontsize{10pt}{6pt}\selectfont
  \textbf{Yuekai Huang}\textsuperscript{\normalfont  1,2,3} \hspace{0.5cm}
  \textbf{Qing Wang}\textsuperscript{1,2,3}  \hspace{0.5cm}
  \textbf{Yihao Huang} \textsuperscript{4} \hspace{0.5cm}
  \textbf{Yang Liu}\textsuperscript{4} \\
  \fontsize{10pt}{6pt}\selectfont
  \textsuperscript{1}State Key Laboratory of Intelligent Game, Beijing, China \\
\fontsize{10pt}{6pt}\selectfont
 \textsuperscript{2}Science and Technology on Integrated Information System Laboratory, \\
  \fontsize{10pt}{6pt}\selectfont
Institute of Software Chinese Academy of Sciences, Beijing, China \\
  \fontsize{10pt}{6pt}\selectfont
  \textsuperscript{3}University of Chinese Academy of Sciences
  \hspace{0.3cm}
  \textsuperscript{4}Nanyang Technological University \\
}
\begin{document}
 \maketitle
\begin{abstract}
Incorporating external knowledge has emerged as a promising way to mitigate outdated knowledge and hallucinations in LLM.
However, external knowledge is often imperfect, encompassing substantial extraneous or even inaccurate content, which interferes with the LLM's utilization of useful knowledge in the context.
This paper seeks to characterize the features of preferred external knowledge and perform empirical studies in imperfect contexts.
Inspired by the chain of evidence (CoE), we characterize that the knowledge preferred by LLMs should maintain both relevance to the question and mutual support among the textual pieces.
Accordingly, we propose a CoE discrimination approach and conduct a comparative analysis between CoE and Non-CoE samples across significance, deceptiveness, and robustness, revealing the LLM's preference for external knowledge that aligns with CoE features.
Furthermore, we selected three representative tasks (RAG-based multi-hop QA, external knowledge poisoning and poisoning defense), along with corresponding SOTA or prevalent baselines. 
By integrating CoE features, the variants achieved significant improvements over the original baselines.

\end{abstract}

\section{Introduction}
\label{sec:introduction}




The parameterized knowledge acquired by large language models (LLMs) through pre-training at a specific point in time becomes outdated with the knowledge evolution or produces hallucination ~\cite{achiam2023gpt,touvron2023llama,anil2023palm2technicalreport}.
Incorporating external knowledge into LLM has emerged as an effective approach to mitigate this problem~\cite{tu2024r,zhao2024retrieval}.
In this context, properties such as the accuracy and reliability of external knowledge are critical for LLMs to provide accurate answers.


However, external knowledge is often imperfect.
In addition to the useful knowledge that users expect LLMs to follow, the context typically contains two types of noise~\cite{chen2024benchmarking,zou2024poisonedrag}:
1) extraneous information, despite showing textual similarities with the question, cannot support the correct answer~\cite{chen2024benchmarking,xiang2024certifiably};
2) inaccurate information, which can mislead LLM to produce incorrect answers~\cite{liu2024open}.
Especially when dealing with complex scenarios such as multi-hop QA, the acquisition of such noise is inevitable due to limitations of retrievers or quality deficiencies in the specialized knowledge bases~\cite{wang2024astuteragovercomingimperfect,dai2024unifying,tang2024multihop}.
This hinders LLMs from effectively using useful knowledge within external contexts and leads to incorrect answers.

Consequently, numerous studies aim to characterize the features of external knowledge that LLMs tend to follow in imperfect contexts (such as confirmation bias, completeness bias, coherent bias, etc.)~\cite{xie2023adaptive,zhang2024evaluating}; or on approaches such as reranking or retrieval to prioritize knowledge with high relevance~\cite{asai2023self,2024Dong}.
However, previous studies primarily suffer from two main deficiencies.
First, while their focus is on qualitative findings, it remains uncertain whether such findings can effectively guide performance improvements in representative tasks~\cite{zhang2024evaluating}.
Second, their research focuses on single-hop QA, in which a single piece of knowledge suffices to answer the question. However, ‌the generalizability of these findings to more complex scenarios (e.g., multi-hop QA) has yet to be confirmed.

In our study, we focus on characterizing what external knowledge is more capable of resisting the surrounding noise and guiding LLMs for better generation.
Inspired by the Chain of Evidence (CoE) theory in criminal procedural law~\cite{murphy2013mismatch}, which requires case-decisive evidence to demonstrate both relevance (pertaining to the case) and interconnectivity (evidence mutually supporting each other) in judicial decisions.
In multi-hop QA, analogously to the scenario where LLMs rely on external knowledge for answering, we consider that the preferred knowledge should show relevance to the question (relevance) and mutual support and complementarity among textual pieces in addressing the question (interconnectivity).
Based on the principle, we first characterize what knowledge can be considered CoE and propose a discrimination approach to determine whether the given external knowledge aligns to the CoE features.
Subsequently, we conduct a comparative analysis of CoE versus Non-CoE samples, examining LLMs' preference for CoE-aligned content across four dimensions below.

\begin{itemize}

\item \textbf{Significance}, we examine whether LLMs demonstrate superior performance when provided with external knowledge exhibiting CoE characteristics, versus cases where the knowledge is relevant but lacks COE characteristics.


\item \textbf{Deceptiveness}, we investigate whether samples that conform to COE characteristics but lead to incorrect answers exhibit higher deceptive potential, effectively inducing LLMs to generate incorrect output.

\item \textbf{Robustness}, we investigate whether the approach effectively mitigates knowledge conflicts and enhances question-answering performance in multi-hop scenarios.

\item \textbf{Usability}, we select three representative tasks (RAG-based multi-hop QA, external knowledge poisoning and defenses) to explore whether CoE can be effectively integrated and enhance the effectiveness of baselines.

 
\end{itemize}


Using HotpotQA~\cite{hotpotqa} and 2WikiMultihopQA~\cite{2wikimultihop} as data sources, we constructed 1,336 multi-hop QA pairs and the corresponding CoE based on the automatic discrimination. 
By applying perturbations to CoE, we also build Non-CoE samples (that is, knowledge lacking the necessary relevance or interconnectivity to establish CoE) for each QA pair.
Subsequently, we conducted a comprehensive evaluation in five state-of-the-art LLMs (GPT-3.5~\cite{openai2022chatgpt}, GPT-4~\cite{achiam2023gpt}, LLama2-13B~\cite{touvron2023llama2}, LLama3-70B~\cite{touvron2023llama}, and Qwen2.5-32B~\cite{qwen2.5_2024}. 

The empirical analysis implies that if external knowledge in the context exhibits CoE characterization, it can better resist interference from extraneous and even inaccurate information and improve multi-hop QA performance.
Building upon these findings, we can effectively enhance existing multi-hop QA approaches and poisoning defenses through performance improvements.
Nevertheless, the observed preference for CoE-compliant external knowledge creates a vulnerability.
Adversaries can deliberately construct false information with CoE characteristics to successfully trick LLMs into generating answers containing factual errors.
Empirically, our investigation uncovers characteristic preferences of LLMs toward external knowledge from both relevance and interconnectivity perspectives, which informs the optimization of knowledge representation and retrieval mechanisms in RAG systems. Practically, our studies demonstrate significant improvements over the SOTA or prevalent baselines in three representative tasks.
The reproduction package is available at: \url{https://anonymous.4open.science/r/ScopeCOE-78D3}.
\section{Related Work}
\label{sec:Background}
In imperfect knowledge augmentation,  there is growing interest in understanding LLMs' knowledge preferences, especially in contexts involving conflicts between external and internal knowledge, as well as contradictions within internal knowledge~\cite{xie2023adaptive,kasai2023realtime, tan2024blinded,jin2024tug, xu2024knowledge, xu2024earth}.

\citet{xie2023adaptive} demonstrated LLMs' bias towards coherent knowledge, revealing that LLMs are highly receptive to external knowledge when presented coherently, even when it conflicts with their parametric knowledge.
\citet{jin2024tug} found that LLMs demonstrate confirmation bias, manifested as their inclination to choose knowledge consistent with their internal memory, regardless of whether it is correct or incorrect.
\citet{chen2022rich} demonstrated LLMs' preference for highly relevant knowledge by manipulating retrieved snippets based on attention scores, showing that LLMs prioritize knowledge with greater relevance to questions.
\citet{zhang2024evaluating} found LLMs perform better when given complete external knowledge, showing completeness bias.


Although existing studies have documented LLMs' knowledge preferences, there exists a significant gap in understanding and measuring the essential features that govern these preferences, especially in complex scenarios like multi-hop QA.
To this end, we manage to characterize and discriminate external knowledge that can help LLMs generate correct responses.

\section{CoE Characterization and Discrimination}

\subsection{CoE Characterization}


Drawing from the law of criminal procedure, judicial decisions in cases require the formation of a CoE through evidence collection~\cite{edmond2011contextual,murphy2013mismatch}.
Such a CoE must demonstrate two properties: relevance (pertaining to the case) and interconnectivity (evidence mutually supporting each other).
In multi-hop QA, the user question is analogous to a legal case, where external contexts constitute the evidentiary collection, and the LLM's answer represents the judicial conclusion drawn through iterative reasoning processes.
Based on this analogy, we hypothesize that in the reasoning process from user query to final answer, LLMs tend to prioritize external knowledge that demonstrates both relevance and interconnectivity.

\begin{figure}[h]
\centering
\setlength{\abovecaptionskip}{5pt}   
  \setlength{\belowcaptionskip}{0pt} 
\includegraphics[width=7.6cm,height=3.3cm]{fig/CoE_inference.png}
\caption{
Example of CoE and the CoE features.
}
\label{fig:CoE_explain}
\end{figure}

Next, we will characterize relevance and interconnectivity from the following three perspectives of textual features presented by external knowledge.
\begin{itemize}
    \item 
    \textbf{Intent} is a noun or noun phrase representing the user's desired answer to their question, and it aims to align the purpose of the user’s question with the ultimate facts derived from external knowledge.
    \item 
    \textbf{Evidence Nodes} are the key entities in a user's question, which imply critical knowledge elements for multi-hop reasoning.
    It ensures logical consistency between the starting and ending points of a single reasoning hop, aligning the user's query with external knowledge.
    \item
    \textbf{Evidence Relations} are logical predicates within the question, indicating the semantic associations between each pair of evidence nodes.
    It is used to verify whether the implicit semantic connections between entities in external knowledge are consistent with the inherent logic in the question.
\end{itemize}

Taking Figure \ref{fig:CoE_explain} as an example, intent specifies ``state location of business'' as the goal, indicating that the user wants to find the state where the business operates. 
The evidence nodes, ``drug stores'', ``CEO'', and ``Warren Bryant'', serve as essential nodes for multi-hop reasoning.
Evidence relations show how these entities are linked, with ``have'' connecting ``drug stores'' to ``CEO'', and ``is'' linking ``CEO'' to ``Warren Bryant''. 
The effectiveness of CoE stems from the synergistic interaction of these three features.
The integration of all three features creates a comprehensive logic chain tailored to the specific question.

\subsection{CoE Discrimination}

Based on the characterized features, we design an approach to discriminate whether external knowledge exhibits CoE features.
Generally, discrimination extracts three CoE features from the user query, i.e., intent, evidence node, and evidence relation. 
These features represent the underlying logic embedded within the user question and serve as objectives for external knowledge alignment.
Subsequently, for a given piece of external knowledge, we verify whether it simultaneously satisfies all three features.
The following introduces the details for the implementation of CoE discrimination.

First, for a user question, we extract its intent, the evidence nodes and the evidence relations using GPT-4o with a hand-crafted prompt.
Second, 
with the extracted CoE features, we discriminate whether external knowledge exhibits them using GPT-4o.
As for intent discrimination, we frame it as a textual entailment task, where external knowledge as a premise and intent as a hypothesis.
We reason whether the hypothesis holds on the basis of the given premise using GPT-4o with prompting.
To discriminate the nodes and relations, we uniformly treat this as a classification task about the ``containment'' logic. 
In implementation, we manually construct distinct prompts for each type of feature, instructing GPT-4o to classify whether an external knowledge contains the extracted nodes and relations.
Finally, external knowledge is considered aligned to the user question only when all required CoE features are present. 
The discrimination prompts are detailed in Appendix \ref{sec:appendix_complete}.

\begin{figure}[h]
\centering
\setlength{\abovecaptionskip}{5pt}   
  \setlength{\belowcaptionskip}{0pt} 
\includegraphics[width=7.8cm,height=4.0cm]{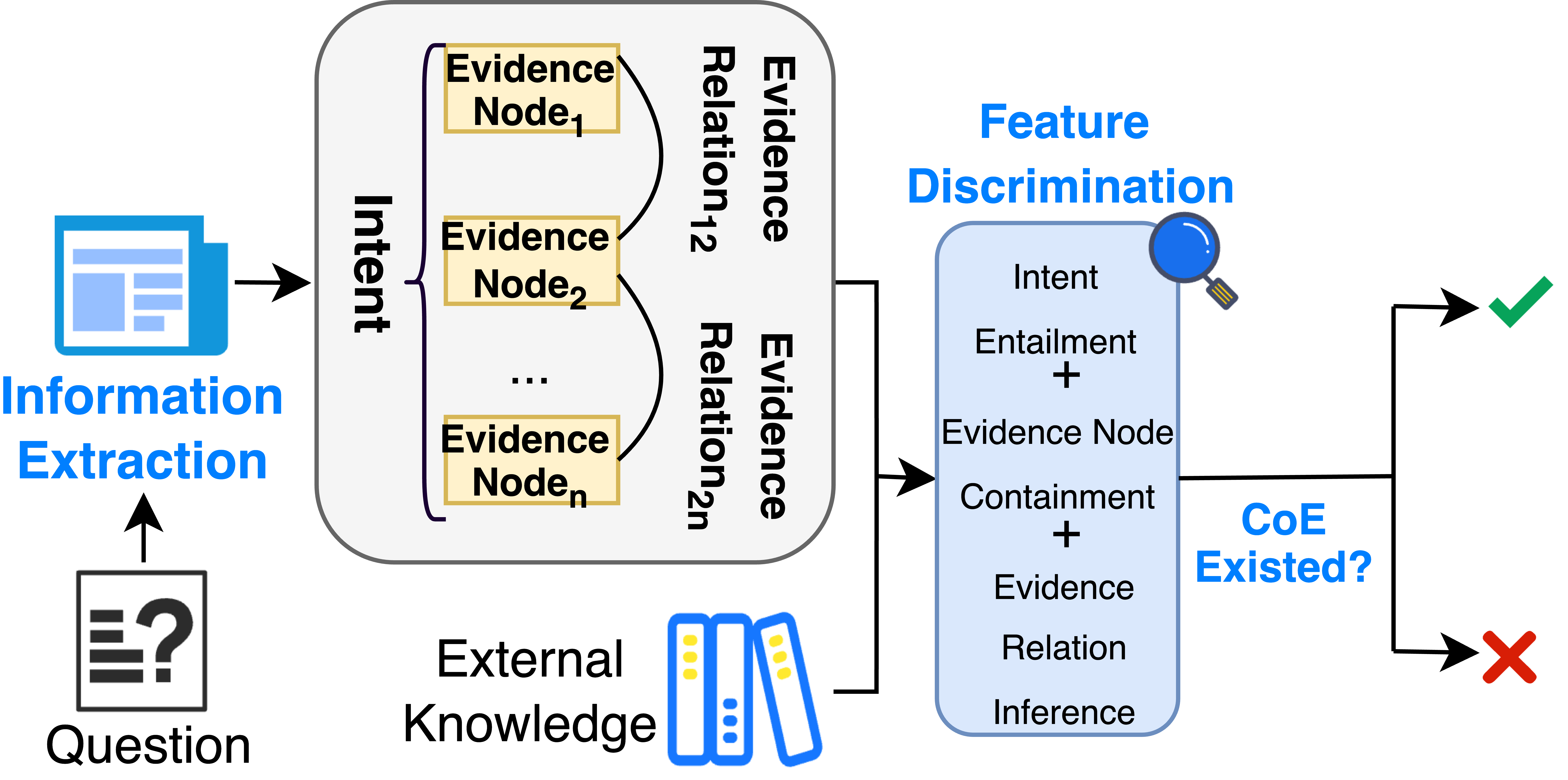}
\caption{
The overview of CoE discrimination.
}
\label{fig:loopjudge}
\end{figure}


\section{Subject Dataset}

\subsection{CoE Knowledge Construction}
\label{sec:dataconstruct}
Oriented to multi-hop QA, we selected two commonly used datasets, HotpotQA and 2WikiMultihopQA as sources.
Both datasets contain samples comprising not just QA pairs, but also the supporting knowledge required to derive each answer.
During construction, the supporting knowledge was specifically designed to capture the necessary logical chains for multi-hop reasoning. 
We consider it to be highly compatible with the COE features, thus qualifying it as a candidate COE knowledge.

Referring to the sample size in previous studies~\cite{jin2024tug,chen2024benchmarking}, 
we randomly sampled 1,000 instances from each dataset and applied the CoE discrimination approach to check whether each candidate exhibits CoE features.
Finally, we obtained 676 and 660 knowledge pieces that contain CoE from candidates, with an average of 4.0 and 3.4 sentences within the supporting knowledge for two datasets, respectively.

\subsection{Non-CoE Knowledge Construction}
\label{sec:non-coe_construct}
Based on the CoE Knowledge, we construct Non-CoE knowledge using two controlled perturbation strategies that are commonly employed in empirical studies of LLM knowledge construction \cite{xie2023adaptive}: sentence-level perturbation (SenP) and word-level perturbation (WordP). 
SenP simulates incomplete knowledge by removing key evidence pieces, while WordP replaces specific evidence nodes with their higher-level expressions.
These strategies ensure fair comparison by maintaining the same question context while only varying the CoE completeness. 
Detailed perturbation procedures and examples are provided in Appendix \ref{sec:appendix_perturbation}.
Subsequently, we construct complete LLM contexts by augmenting COE and Non-COE knowledge with extraneous and inaccurate information.
Through comparative analysis, we systematically examine their performance differences across various scenarios.

\section{Significance Assessment}
\label{sec:effectiveness}
In practice, the context of LLMs often contains non-trivial noise due to limitations in knowledge base quality or retriever performance \cite{liu2024open}. 
This results in situations where even when useful knowledge is retrieved, other noise may interfere with it, ultimately preventing LLMs from generating desired content.
In this section, we aim to investigate whether knowledge that aligns with CoE features can more effectively resist noise in the context and help LLMs answer multi-hop questions.
To this end, we employ a comparative analysis methodology.
We retrieve text fragments that are lexically similar but offer limited relevance to QA, simulating potential extraneous information interference in contexts. 
These fragments are then combined with either CoE or Non-CoE knowledge to form complete contexts, followed by an analysis of whether significant performance differences exist between the two groups.

\subsection{Experimental Design}
\label{sec:evaACC}

We design a comprehensive experimental framework to evaluate the effectiveness of CoE and Non-CoE ($\text{Non-CoE}_{SenP}$ and $\text{Non-CoE}_{WordP}$) under different noise conditions.
Our analysis focuses on four key dimensions: (1) the performance comparison between CoE and Non-CoE on LLMs, (2) the impact of varying extraneous ratios on their effectiveness, and (3) the performance of CoE in single-hop and multi-hop scenarios.
(4) the influence of different CoE features on LLM performance.
We inject extraneous information at four different ratios (from 0 to 0.75, with 0.25 intervals) to examine how each approach maintains its effectiveness.

For evaluation, we select five representative LLMs that span both closed-source (GPT-3.5, GPT-4) and open-source LLMs (LLama2-13B, LLama3-70B, Qwen2.5-32B) to ensure comprehensive coverage across different model scales and architectures. Following general QA evaluation protocols \citet{adlakha2024}, we use GPT-4 as the judge to compute the accuracy (ACC) between model outputs and ground truth answers.
To understand CoE's significance more comprehensively, we conduct experiments on a single-hop dataset and perform ablation studies by removing different CoE features. Detailed experimental settings are provided in Appendix \ref{sec:appendix_ablation} and \ref{sec:single-hop}.

\subsection{Results and Discussion}
\begin{table}[t]\huge
  \caption{
  LLMs' Accuracy (ACC) on CoE and Non-CoE.
  }
  \label{tab:RQ1_performance}
  
\resizebox{0.5\textwidth }{!}{
\begin{threeparttable}
\begin{tabular}{cc|ccccccc}
\toprule
\multirow{3}{*}{\textbf{Model}} &
  \multirow{3}{*}{\textbf{\shortstack{Irrelevant\\ Proportion}}} & 

  \multicolumn{3}{c}{\textbf{HotpotQA}} & &
  \multicolumn{3}{c}{\textbf{2WikiMultihopQA}}
  \\    \cline{3-5} \cline{7-9}

   &

   &

  \multirow{2}{*}{\textbf{CoE}} &
  \multicolumn{2}{c}{\textbf{Non-CoE}} & 
  &
  \multirow{2}{*}{\textbf{CoE}} &
  \multicolumn{2}{c}{\textbf{Non-CoE}}
\\
\cline{4-5} \cline{8-9}

   &

   &

 &
  \multicolumn{1}{c}{\textit{\textbf{SenP}}} & 
  \multicolumn{1}{c}{\textit{\textbf{WordP}}} &
  & &
  \multicolumn{1}{c}{\textit{\textbf{SenP}}} &
  \multicolumn{1}{c}{\textit{\textbf{WordP}}}
  
  \\ \midrule
\multirow{4}{*}{{GPT-3.5}} &
  \textit{0} &
  
  \multicolumn{1}{c}{\textbf{91.9\%}} &
  \multicolumn{1}{c}{77.9\%$^*$} &
  \multicolumn{1}{c}{79.1\%$^*$} &
  &
  \multicolumn{1}{c}{\textbf{97.4\%}} &
  \multicolumn{1}{c}{74.1\%$^*$} &
  \multicolumn{1}{c}{83.5\%$^*$} 
  
  \\ 

  &
  \textit{0.25} &
 
  \multicolumn{1}{c}{\textbf{90.3\%}} &
  \multicolumn{1}{c}{75.6\%$^*$} &
  \multicolumn{1}{c}{77.5\%$^*$} &
  &
  \multicolumn{1}{c}{\textbf{96.9\%}} &
  \multicolumn{1}{c}{68.2\%$^*$} &
  \multicolumn{1}{c}{81.2\%$^*$} 
  \\

  &
  \textit{0.5} &

  \multicolumn{1}{c}{\textbf{89.9\%}} &
  \multicolumn{1}{c}{73.1\%$^*$} &
  \multicolumn{1}{c}{75.4\%$^*$} &
  &
   \multicolumn{1}{c}{\textbf{96.5\%}} &
  \multicolumn{1}{c}{66.4\%$^*$} &
  \multicolumn{1}{c}{82.6\%$^*$} 
  \\

  &
  \textit{0.75} &

  \multicolumn{1}{c}{\textbf{88.9\%}} &
  \multicolumn{1}{c}{65.7\%$^*$} &
  \multicolumn{1}{c}{74.5\%$^*$} &
  &
   \multicolumn{1}{c}{\textbf{95.4\%}} &
  \multicolumn{1}{c}{58.4\%$^*$} &
  \multicolumn{1}{c}{70.8\%$^*$} 
  \\ 
  
   \midrule

\multirow{4}{*}{{GPT-4}} &
  \textit{0} &
  
  \multicolumn{1}{c}{\textbf{93.5\%}} &
  \multicolumn{1}{c}{83.4\%$^*$} &
  \multicolumn{1}{c}{86.4\%$^*$} &
  &
  \multicolumn{1}{c}{\textbf{93.7\%}} &
  \multicolumn{1}{c}{67.7\%$^*$} &
  \multicolumn{1}{c}{79.4\%$^*$} 
  
  \\ 

  &
  \textit{0.25} &
 
  \multicolumn{1}{c}{\textbf{93.4\%}} &
  \multicolumn{1}{c}{82.3\%$^*$} &
  \multicolumn{1}{c}{86.4\%$^*$} &
  &
  \multicolumn{1}{c}{\textbf{94.0\%}} &
  \multicolumn{1}{c}{70.9\%$^*$} &
  \multicolumn{1}{c}{80.1\%$^*$} 
  \\

  &
  \textit{0.5} &

  \multicolumn{1}{c}{\textbf{91.8\%}} &
  \multicolumn{1}{c}{82.0\%$^*$} &
  \multicolumn{1}{c}{86.5\%$^*$} &
  &
   \multicolumn{1}{c}{\textbf{95.4\%}} &
  \multicolumn{1}{c}{71.5\%$^*$}  &
  \multicolumn{1}{c}{77.3\%$^*$} 
  \\

  &
  \textit{0.75} &

  \multicolumn{1}{c}{\textbf{91.2\%}} &
  \multicolumn{1}{c}{80.1\%$^*$} &
  \multicolumn{1}{c}{83.8\%$^*$} &
  &
   \multicolumn{1}{c}{\textbf{95.9\%}} &
  \multicolumn{1}{c}{64.9\%$^*$} &
  \multicolumn{1}{c}{74.4\%$^*$} 
  \\ 

   \midrule
   
\multirow{4}{*}{{Llama2-13B}} &
  \textit{0} &
  
  \multicolumn{1}{c}{\textbf{89.9\%}} &
  \multicolumn{1}{c}{87.1\%$^*$} &
  \multicolumn{1}{c}{88.8\%$^*$} &
  &
  \multicolumn{1}{c}{\textbf{96.5\%}} &
  \multicolumn{1}{c}{95.3\%$^*$} &
  \multicolumn{1}{c}{93.3\%$^*$} 
  
  \\ 

  &
  \textit{0.25} &
 
  \multicolumn{1}{c}{\textbf{87.9\%}} &
  \multicolumn{1}{c}{84.2\%$^*$} &
  \multicolumn{1}{c}{85.2\%$^*$} &
  &
  \multicolumn{1}{c}{\textbf{95.9\%}} &
  \multicolumn{1}{c}{93.7\%$^*$} &
  \multicolumn{1}{c}{91.9\%$^*$} 
  \\

  &
  \textit{0.5} &

  \multicolumn{1}{c}{\textbf{86.4\%}} &
  \multicolumn{1}{c}{82.8\%$^*$} &
  \multicolumn{1}{c}{84.0\%$^*$} &
  &
   \multicolumn{1}{c}{\textbf{93.8\%}} &
  \multicolumn{1}{c}{91.2\%$^*$}  &
  \multicolumn{1}{c}{90.0\%$^*$} \\

  &
  \textit{0.75} &

  \multicolumn{1}{c}{\textbf{85.8\%}} &
  \multicolumn{1}{c}{79.5\%$^*$} &
  \multicolumn{1}{c}{82.9\%$^*$} &
  &
   \multicolumn{1}{c}{\textbf{90.9\%}} &
  \multicolumn{1}{c}{86.6\%$^*$} &
  \multicolumn{1}{c}{86.3\%$^*$} 
  \\ 

   \midrule
\multirow{4}{*}{{Llama3-70B}} &
\textit{0} &
  
  \multicolumn{1}{c}{\textbf{92.5\%}} &
  \multicolumn{1}{c}{76.8\%$^*$} &
  \multicolumn{1}{c}{74.5\%$^*$} &
  &
  \multicolumn{1}{c}{\textbf{95.7\%}} &
  \multicolumn{1}{c}{79.5\%$^*$} &
  \multicolumn{1}{c}{73.3\%$^*$} 
  
  \\ 

  &
  \textit{0.25} &
 
  \multicolumn{1}{c}{\textbf{92.9\%}} &
  \multicolumn{1}{c}{74.1\%$^*$} &
  \multicolumn{1}{c}{76.1\%$^*$} &
  &
  \multicolumn{1}{c}{\textbf{93.7\%}} &
  \multicolumn{1}{c}{80.3\%$^*$} &
  \multicolumn{1}{c}{71.4\%$^*$} 
  \\

  &
  \textit{0.5} &

  \multicolumn{1}{c}{\textbf{91.1\%}} &
  \multicolumn{1}{c}{72.6\%$^*$} &
  \multicolumn{1}{c}{76.8\%$^*$} &
  &
   \multicolumn{1}{c}{\textbf{95.9\%}} &
  \multicolumn{1}{c}{76.7\%$^*$} &
  \multicolumn{1}{c}{69.6\%$^*$} \\

  &
  \textit{0.75} &

  \multicolumn{1}{c}{\textbf{90.5\%}} &
  \multicolumn{1}{c}{69.8\%$^*$} &
  \multicolumn{1}{c}{68.3\%$^*$} &
  &
   \multicolumn{1}{c}{\textbf{93.1\%}} &
  \multicolumn{1}{c}{72.3\%$^*$} &
  \multicolumn{1}{c}{67.3\%$^*$} 
  
   \\
   \midrule
\multirow{4}{*}{{Qwen2.5-32B}} &
\textit{0} &
  
  \multicolumn{1}{c}{\textbf{87.8\%}} &
  \multicolumn{1}{c}{71.3\%$^*$} &
  \multicolumn{1}{c}{75.7\%$^*$} &
  &
  \multicolumn{1}{c}{\textbf{90.7\%}} &
  \multicolumn{1}{c}{53.1\%$^*$} &
  \multicolumn{1}{c}{67.0\%$^*$} 
  
  \\ 

  &
  \textit{0.25} &
 
  \multicolumn{1}{c}{\textbf{87.2\%}} &
  \multicolumn{1}{c}{38.6\%$^*$} &
  \multicolumn{1}{c}{64.9\%$^*$} &
  &
  \multicolumn{1}{c}{\textbf{91.3\%}} &
  \multicolumn{1}{c}{29.5\%$^*$} &
  \multicolumn{1}{c}{49.4\%$^*$} 
  \\

  &
  \textit{0.5} &

  \multicolumn{1}{c}{\textbf{86.1\%}} &
  \multicolumn{1}{c}{37.7\%$^*$} &
  \multicolumn{1}{c}{64.3\%$^*$} &
  &
   \multicolumn{1}{c}{\textbf{92.1\%}} &
  \multicolumn{1}{c}{27.8\%$^*$} &
  \multicolumn{1}{c}{47.5\%$^*$} \\

  &
  \textit{0.75} &

  \multicolumn{1}{c}{\textbf{88.0\%}} &
  \multicolumn{1}{c}{37.3\%$^*$} &
  \multicolumn{1}{c}{57.2\%$^*$} &
  &
   \multicolumn{1}{c}{\textbf{91.9\%}} &
  \multicolumn{1}{c}{22.2\%$^*$} &
  \multicolumn{1}{c}{45.9\%$^*$} \\
   
   \bottomrule
\end{tabular}
 \begin{tablenotes}
\item[*] indicates statistical significance compared to CoE (p < 0.05)
\end{tablenotes}
        \end{threeparttable}}
       
\end{table}
Table \ref{tab:RQ1_performance} shows the ACC in different ratios of extraneous information.
Comparing the CoE and Non-CoE groups, the results show that CoE achieves an average ACC of 92.0\% across five LLMs and two datasets, outperforming $\text{Non-CoE}_{SenP}$ and $\text{Non-CoE}_{WordP}$ by 22.5\% and 16.3\%, respectively. 
This substantial improvement suggests that external knowledge that exhibiting CoE features enables LLMs to utilize it and achieve better performance. 


For the proportion of extraneous information in the context, as the ratio increases from 0\% to 75\%, CoE's ACC only decreases by 1.8\%, while the ACC of Non-CoE variants decreases more significantly: 12.9\% for $\text{Non-CoE}_{SenP}$ and 9.0\% for $\text{Non-CoE}_{WordP}$. 
This robustness suggests that external knowledge that exhibits CoE features helps LLMs maintain consistent comprehension and reasoning faced with ‌noise of different magnitudes.


Analyzing the impact of reasoning complexity, experiments show that single-hop reasoning maintains the most stable performance (>92.0\% ACC) under increasing extraneous information, followed by three-hop reasoning (>90.0\% ACC), while two-hop reasoning exhibits higher sensitivity, with ACC dropping from 91.0\% to 88.0\%. This pattern suggests that CoE is particularly effective in simpler reasoning scenarios, while maintaining strong performance in more complex cases. Detailed experimental results and analysis are provided in Appendix \ref{sec:single-hop}.

In addition, ablation studies further demonstrate the varying impacts of CoE features: removing intent causes the largest accuracy drop (33.9\%), followed by evidence nodes (13.6\%), while evidence relation removal has the smallest impact (10.7\%). 
It indicates that explicit reasoning intent is crucial for guiding LLMs' responses. Detailed experimental results and analysis are provided in Appendix \ref{sec:appendix_ablation}.

\textbf{Summary:} If external knowledge exhibits CoE characterization, it can better resist interference from extraneous information and improve multi-hop QA performance.
Moreover, LLMs exhibit greater resistance if there exists external knowledge exhibiting CoE features, as the proportion of extraneous information increases.
For practical guidance, optimizing the retriever to prioritize knowledge exhibiting CoE features can effectively enhance performance of multi-hop QA.

\section{Deceptiveness Assessment}
\label{sec:faithfulness}

Given that CoE represents structured reasoning chains, it is crucial to examine whether such well-formed evidence paths could amplify the deceptive effect of poisoned knowledge. Therefore, we investigate a more challenging scenario, where the CoE contains factual errors, to determine whether LLMs can still be effectively misled and produce answers consistent with the incorrect information embedded in the CoE.


\subsection{Experimental Design}
\label{sec_wrongans}
To investigate the deceptiveness of incorrect external knowledge under imperfect conditions, we design a comprehensive evaluation framework comparing CoE and Non-CoE.
Our analysis focuses on three key dimensions: (1) the comparative effectiveness between CoE and Non-CoE ($\text{Non-CoE}_{SenP}$ and $\text{Non-CoE}_{WordP}$) in misleading LLMs with incorrect information, and (2) how their deceptive capabilities change under varying ratios of irrelevant information, and (3) the influence of different CoE features on LLM deception effectiveness.

In constructing incorrect information for both CoE and Non-CoE, we carefully preserve the semantic type and format of original answers (e.g., replacing "United States" with "Canada" while maintaining consistent structures) to ensure fair comparison. 
Following Section \ref{sec:evaACC}, we inject irrelevant information at the same ratios to examine how each approach maintains its deceptive effectiveness.

Using the same subject LLMs as Section \ref{sec:evaACC} and following standard evaluation protocols \citet{adlakha2024}, we use GPT-4o as the judge to compute the Attack Success Rate (ASR), defined as the proportion of successfully misled LLM outputs.
We also conduct ablation studies to analyze how different CoE features midleading LLM responses. Detailed experimental settings are provided in \ref{sec:appendix_ablation}.

\subsection{Results and Discussion}
\begin{table}[t]\huge
  \caption{Attack Success Rate (ASR) of CoE and Non-CoE against LLMs.}
  \label{tab:RQ2_performance}
  
\resizebox{0.5\textwidth }{!}{
\begin{threeparttable}
\begin{tabular}{cc|ccccccc}
\toprule
\multirow{3}{*}{\textbf{Model}} &
  \multirow{3}{*}{\textbf{\shortstack{Irrelevant\\ Proportion}}} & 

  \multicolumn{3}{c}{\textbf{HotpotQA}} & &
  \multicolumn{3}{c}{\textbf{2WikiMultihopQA}}
  \\    \cline{3-5} \cline{7-9}

   &

   &

  \multirow{2}{*}{\textbf{CoE}} &
  \multicolumn{2}{c}{\textbf{Non-CoE}} & 
  &
  \multirow{2}{*}{\textbf{CoE}} &
  \multicolumn{2}{c}{\textbf{Non-CoE}}
\\
\cline{4-5} \cline{8-9}

   &

   &

 &
  \multicolumn{1}{c}{\textit{\textbf{SenP}}} & 
  \multicolumn{1}{c}{\textit{\textbf{WordP}}} &
  & &
  \multicolumn{1}{c}{\textit{\textbf{SenP}}} &
  \multicolumn{1}{c}{\textit{\textbf{WordP}}}
  \\ \midrule
\multirow{4}{*}{{GPT-3.5}} &
  \textit{0} &
  
  \multicolumn{1}{c}{\textbf{86.1\%}} &
  \multicolumn{1}{c}{75.6\%$^*$} &
  \multicolumn{1}{c}{83.1\%$^*$} &
  &
  \multicolumn{1}{c}{\textbf{85.0\%}} &
  \multicolumn{1}{c}{58.5\%$^*$} &
  \multicolumn{1}{c}{57.4\%$^*$} 
  
  \\ 

  &
  \textit{0.25} &
 
  \multicolumn{1}{c}{\textbf{85.8\%}} &
  \multicolumn{1}{c}{76.0\%$^*$} &
   \multicolumn{1}{c}{79.1\%$^*$} &
  &
  \multicolumn{1}{c}{\textbf{86.5\%}} &
  \multicolumn{1}{c}{53.8\%$^*$} &
   \multicolumn{1}{c}{52.4\%$^*$} 
  \\

  &
  \textit{0.5} &

  \multicolumn{1}{c}{\textbf{84.7\%}} &
  \multicolumn{1}{c}{72.2\%$^*$} &
  \multicolumn{1}{c}{77.8\%$^*$} &
  &
   \multicolumn{1}{c}{\textbf{84.2\%}} &
  \multicolumn{1}{c}{50.0\%$^*$} &
  \multicolumn{1}{c}{48.8\%$^*$}
  \\

  &
  \textit{0.75} &

  \multicolumn{1}{c}{\textbf{78.4\%}} &
  \multicolumn{1}{c}{72.0\%$^*$} &
  \multicolumn{1}{c}{73.7\%$^*$} &
  &
   \multicolumn{1}{c}{\textbf{83.3\%}} &
  \multicolumn{1}{c}{45.2\%$^*$} &
  \multicolumn{1}{c}{44.9\%$^*$}
  \\ 
  
   \midrule
   \multirow{4}{*}{{GPT-4}} &
  \textit{0} &
  
  \multicolumn{1}{c}{\textbf{86.5\%}} &
  \multicolumn{1}{c}{52.2\%$^*$} &
  \multicolumn{1}{c}{59.0\%$^*$} &
  &
  \multicolumn{1}{c}{\textbf{85.4\%}} &
  \multicolumn{1}{c}{68.8\%$^*$} &
  \multicolumn{1}{c}{76.2\%$^*$} 
  
  \\ 

  &
  \textit{0.25} &
 
  \multicolumn{1}{c}{\textbf{85.5\%}} &
  \multicolumn{1}{c}{50.5\%$^*$} &
   \multicolumn{1}{c}{58.9\%$^*$} &
  &
  \multicolumn{1}{c}{\textbf{87.2\%}} &
  \multicolumn{1}{c}{67.0\%$^*$} &
   \multicolumn{1}{c}{73.2\%$^*$} 
  \\

  &
  \textit{0.5} &

  \multicolumn{1}{c}{\textbf{84.0\%}} &
  \multicolumn{1}{c}{46.8\%$^*$} &
  \multicolumn{1}{c}{52.7\%$^*$} &
  &
   \multicolumn{1}{c}{\textbf{90.6\%}} &
  \multicolumn{1}{c}{65.2\%$^*$} &
  \multicolumn{1}{c}{76.8\%$^*$}
  \\

  &
  \textit{0.75} &

  \multicolumn{1}{c}{\textbf{78.2\%}} &
  \multicolumn{1}{c}{43.2\%$^*$} &
  \multicolumn{1}{c}{50.5\%$^*$} &
  &
   \multicolumn{1}{c}{\textbf{92.7\%}} &
  \multicolumn{1}{c}{62.3\%$^*$} &
  \multicolumn{1}{c}{75.1\%$^*$}
  \\ 
  
   \midrule
\multirow{4}{*}{{Llama2-13B}} &
  \textit{0} &
  
  \multicolumn{1}{c}{\textbf{78.2\%}} &
  \multicolumn{1}{c}{76.9\%$^*$} &
  \multicolumn{1}{c}{72.9\%$^*$} &
  &
  \multicolumn{1}{c}{\textbf{91.5\%}} &
  \multicolumn{1}{c}{89.8\%$^*$} &
  \multicolumn{1}{c}{88.6\%$^*$} 
  
  \\ 

  &
  \textit{0.25} &
 
  \multicolumn{1}{c}{\textbf{77.1\%}} &
  \multicolumn{1}{c}{74.1\%$^*$} &
   \multicolumn{1}{c}{67.3\%$^*$} &
  &
  \multicolumn{1}{c}{\textbf{89.8\%}} &
  \multicolumn{1}{c}{87.5\%$^*$} &
  \multicolumn{1}{c}{86.3\%$^*$}
  \\

  &
  \textit{0.5} &
\multicolumn{1}{c}{\textbf{71.6\%}} &
  \multicolumn{1}{c}{70.0\%$^*$} &
  \multicolumn{1}{c}{67.5\%$^*$} &
  &
   \multicolumn{1}{c}{\textbf{89.1\%}} &
  \multicolumn{1}{c}{86.8\%$^*$} &
  \multicolumn{1}{c}{85.1\%$^*$}
  \\

  &
  \textit{0.75} &

  \multicolumn{1}{c}{\textbf{69.1\%}} &
  \multicolumn{1}{c}{64.5\%$^*$} &
   \multicolumn{1}{c}{64.8\%$^*$} &
  &
   \multicolumn{1}{c}{\textbf{84.1\%}} &
  \multicolumn{1}{c}{81.6\%$^*$} &
  \multicolumn{1}{c}{82.1\%$^*$}
  \\ 

   \midrule
\multirow{4}{*}{{Llama3-70B}} &
\textit{0} &
  
  \multicolumn{1}{c}{\textbf{82.8\%}} &
  \multicolumn{1}{c}{76.9\%$^*$} &
  \multicolumn{1}{c}{72.8\%$^*$} &
  &
  \multicolumn{1}{c}{\textbf{89.7\%}} &
  \multicolumn{1}{c}{77.1\%$^*$} &
  \multicolumn{1}{c}{72.1\%$^*$}
  
  \\ 

  &
  \textit{0.25} &
 
  \multicolumn{1}{c}{\textbf{81.6\%}} &
  \multicolumn{1}{c}{75.1\%$^*$} &
   \multicolumn{1}{c}{71.9\%$^*$} &
  &
  \multicolumn{1}{c}{\textbf{89.5\%}} &
  \multicolumn{1}{c}{72.1\%$^*$} &
   \multicolumn{1}{c}{70.4\%$^*$}
  \\

  &
  \textit{0.5} &

  \multicolumn{1}{c}{\textbf{78.0\%}} &
  \multicolumn{1}{c}{71.7\%$^*$} &
    \multicolumn{1}{c}{68.0\%$^*$} &
  &
   \multicolumn{1}{c}{\textbf{88.9\%}} &
  \multicolumn{1}{c}{69.4\%$^*$} &
  \multicolumn{1}{c}{66.5\%$^*$}
  \\

  &
  \textit{0.75} &

  \multicolumn{1}{c}{\textbf{78.2\%}} &
  \multicolumn{1}{c}{62.9\%$^*$} &
   \multicolumn{1}{c}{64.1\%$^*$} &
  &
   \multicolumn{1}{c}{\textbf{89.8\%}} &
  \multicolumn{1}{c}{51.4\%$^*$} &
  \multicolumn{1}{c}{53.7\%$^*$}
  
   \\
   
    \midrule
\multirow{4}{*}{{Qwen2.5-32B}} &
\textit{0} &
  
  \multicolumn{1}{c}{\textbf{90.6\%}} &
  \multicolumn{1}{c}{68.9\%$^*$} &
  \multicolumn{1}{c}{79.1\%$^*$} &
  &
  \multicolumn{1}{c}{\textbf{93.7\%}} &
  \multicolumn{1}{c}{43.5\%$^*$} &
  \multicolumn{1}{c}{65.8\%$^*$}
  
  \\ 

  &
  \textit{0.25} &
 
  \multicolumn{1}{c}{\textbf{87.7\%}} &
  \multicolumn{1}{c}{67.3\%$^*$} &
   \multicolumn{1}{c}{80.0\%$^*$} &
  &
  \multicolumn{1}{c}{\textbf{93.6\%}} &
  \multicolumn{1}{c}{47.2\%$^*$} &
   \multicolumn{1}{c}{67.3\%$^*$}
  \\

  &
  \textit{0.5} &

  \multicolumn{1}{c}{\textbf{86.3\%}} &
  \multicolumn{1}{c}{64.1\%$^*$} &
    \multicolumn{1}{c}{76.5\%$^*$} &
  &
   \multicolumn{1}{c}{\textbf{93.1\%}} &
  \multicolumn{1}{c}{47.0\%$^*$} &
  \multicolumn{1}{c}{68.6\%$^*$}
  \\

  &
  \textit{0.75} &

  \multicolumn{1}{c}{\textbf{85.8\%}} &
  \multicolumn{1}{c}{62.9\%$^*$} &
   \multicolumn{1}{c}{74.2\%$^*$} &
  &
   \multicolumn{1}{c}{\textbf{94.0\%}} &
  \multicolumn{1}{c}{46.5\%$^*$} &
  \multicolumn{1}{c}{65.6\%$^*$}
  \\
   \bottomrule
\end{tabular}
\begin{tablenotes}
\item[*] indicates statistical significance compared to CoE (p < 0.05)
\end{tablenotes}
        \end{threeparttable}
}
\end{table}
Table 3 shows the ASR of LLMs with external knowledge under CoE and two types of Non-CoE samples leading to incorrect answers. 
The results show that the average ASR reaches 85.4\% for the COE group, which is 20.6\% and 16.2\% higher than $\text{Non-CoE}_{SenP}$ and $\text{Non-CoE}_{WordP}$, respectively. 
The results imply that CoE demonstrates significant deception in misleading LLMs when it contains factual errors.

Combining with the ratio of the irrelevant information,
as the ratio increases from 0\% to 75\%, CoE's ASR only decreases by 3.6\%, while the attack effectiveness of Non-CoE variants drops more significantly (9.7\% for $\text{Non-CoE}_{SenP}$ and 7.9\% for $\text{Non-CoE}_{WordP}$).
In general, CoE's attack effectiveness remains more stable against LLMs when faced with irrelevant noise variations, outperforming two types of Non-CoE samples.
Another noteworthy finding is that when the CoE leads to an incorrect answer, its performance metrics are on average 6.6\% lower than those of a correct CoE (Table \ref{tab:RQ1_performance}).
This phenomenon probably stems from the parametric knowledge inherent in LLMs, which confers a degree of resistance to poisoned or erroneous knowledge input.

Ablation studies reveal the varying impacts of CoE components on LLM deception: removing evidence nodes leads to the highest ASR (78.4\%), followed by removing evidence relations (64.4\%) and intent (53.6\%). However, the absence of evidence nodes results in reduced stability against irrelevant knowledge, with ASR dropping by 9.4\%, indicating their vital role in maintaining deceptive effectiveness under noisy scenarios. Detailed analysis is provided in Appendix \ref{sec:appendix_ablation}.

\textbf{Summary:} External knowledge exhibiting CoE characteristics demonstrates significant deceptiveness in misleading LLMs when it contains factual errors.
This implies that external knowledge matching CoE features requires elevated prioritized safeguards due to their potent deceptiveness.

\section{Robustness Assessment}
\label{sec:robustness}


In addition to examining CoE's performance when confronted with irrelevant information, we further investigate its resilience against knowledge conflicts in the context, e.g., cases where adversarial attacks have compromised other retrieved contexts.


\subsection{Experimental Design}
Starting from the CoE and Non-CoE samples, we continuously inject conflicting knowledge into their context at varying ratios to simulate scenarios where incorrect knowledge is retrieved or context is poisoned. 
We then observe and compare the performance of both sample groups in multi-hop QA to analyze whether CoE samples exhibit stronger robustness to misinformation interference. 


First, we construct conflicting knowledge through two strategies: (1) replacing correct statements with contradictory ones in CoE/Non-CoE sentences, and (2) using GPT-4o to generate diverse contradictory expressions following previous work~\cite{ChenZC22,zhou2023context,jin2024tug}. 
We then inject these contradictory statements and gradually increase their proportion (from 0 to 0.75, with 0.25 intervals) in the contexts.
After that, we obtain the answers from LLMs with the question and the constructed CoE and Non-CoE contexts.
Following standard evaluation protocols \citet{adlakha2024} for multi-hop QA, we use GPT-4o as the evaluator and compute ACC.
We also conduct ablation studies to analyze how different CoE features affect conflict handling capabilities. 
Detailed experimental settings are provided in Appendix \ref{sec:appendix_ablation}.





\subsection{Results and Discussion}
\begin{table}[t]\huge
  \caption{LLMs' Accuracy (ACC) with CoE and Non-CoE surrounded by misinformation.}
  \label{tab:RQ3_robust}
  
\resizebox{0.5\textwidth }{!}{
\begin{threeparttable}
\begin{tabular}{cc|ccccccc}
\toprule
\multirow{3}{*}{\textbf{Model}} &
  \multirow{3}{*}{\textbf{\shortstack{Misinformation\\ Proportion}}} & 

  \multicolumn{3}{c}{\textbf{HotpotQA}} & &
  \multicolumn{3}{c}{\textbf{2WikiMultihopQA}}
  \\    \cline{3-5} \cline{7-9}

   &

   &

  \multirow{2}{*}{\textbf{CoE}} &
  \multicolumn{2}{c}{\textbf{Non-CoE}} & 
  &
  \multirow{2}{*}{\textbf{CoE}} &
  \multicolumn{2}{c}{\textbf{Non-CoE}}
\\
\cline{4-5} \cline{8-9}

   &

   &

 &
  \multicolumn{1}{c}{\textit{\textbf{SenP}}} & 
  \multicolumn{1}{c}{\textit{\textbf{WordP}}} &
  & &
  \multicolumn{1}{c}{\textit{\textbf{SenP}}} &
  \multicolumn{1}{c}{\textit{\textbf{WordP}}}
  \\ \midrule
\multirow{4}{*}{{GPT-3.5}} &
  \textit{0} &
  
  \multicolumn{1}{c}{\textbf{91.9\%}} &
  \multicolumn{1}{c}{77.9\%$^*$} &
  \multicolumn{1}{c}{79.1\%$^*$} &
  &
  \multicolumn{1}{c}{\textbf{97.4\%}} &
  \multicolumn{1}{c}{74.1\%$^*$} &
  \multicolumn{1}{c}{83.5\%$^*$}
  
  \\ 

  &
  \textit{0.25} &
 
  \multicolumn{1}{c}{\textbf{81.8\%}} &
  \multicolumn{1}{c}{62.5\%$^*$} &
  \multicolumn{1}{c}{64.0\%$^*$} &
  &
  \multicolumn{1}{c}{\textbf{85.3\%}} &
  \multicolumn{1}{c}{40.6\%$^*$} &
  \multicolumn{1}{c}{63.8\%$^*$}
  \\

  &
  \textit{0.5} &

  \multicolumn{1}{c}{\textbf{82.0\%}} &
  \multicolumn{1}{c}{63.0\%$^*$} &
  \multicolumn{1}{c}{65.7\%$^*$} &
  &
  \multicolumn{1}{c}{\textbf{65.5\%}} &
  \multicolumn{1}{c}{43.4\%$^*$} &
  \multicolumn{1}{c}{52.3\%$^*$} \\

  &
  \textit{0.75} &

  \multicolumn{1}{c}{\textbf{75.7\%}} &
  \multicolumn{1}{c}{58.9\%$^*$} &
  \multicolumn{1}{c}{60.8\%$^*$} &
  &
  \multicolumn{1}{c}{\textbf{55.5\%}} &
  \multicolumn{1}{c}{29.8\%$^*$} &
  \multicolumn{1}{c}{30.4\%$^*$}
  \\ 
  
   \midrule
\multirow{4}{*}{{GPT-4}} &
  \textit{0} &
  
  \multicolumn{1}{c}{\textbf{93.5\%}} &
  \multicolumn{1}{c}{83.4\%$^*$} &
  \multicolumn{1}{c}{86.4\%$^*$} &
  &
  \multicolumn{1}{c}{\textbf{93.7\%}} &
  \multicolumn{1}{c}{67.7\%$^*$} &
  \multicolumn{1}{c}{79.4\%$^*$}
  
  \\ 

  &
  \textit{0.25} &
 
  \multicolumn{1}{c}{\textbf{95.3\%}} &
  \multicolumn{1}{c}{89.7\%$^*$} &
  \multicolumn{1}{c}{89.9\%$^*$} &
  &
  \multicolumn{1}{c}{\textbf{96.5\%}} &
  \multicolumn{1}{c}{86.0\%$^*$} &
  \multicolumn{1}{c}{91.9\%$^*$} 
  \\

  &
  \textit{0.5} &

  \multicolumn{1}{c}{\textbf{90.7\%}} &
  \multicolumn{1}{c}{84.6\%$^*$} &
  \multicolumn{1}{c}{87.4\%$^*$} &
  &
   \multicolumn{1}{c}{\textbf{90.7\%}} &
  \multicolumn{1}{c}{78.3\%$^*$}  &
  \multicolumn{1}{c}{84.2\%$^*$} \\

  &
  \textit{0.75} &

  \multicolumn{1}{c}{\textbf{86.6\%}} &
  \multicolumn{1}{c}{75.2\%$^*$} &
  \multicolumn{1}{c}{78.1\%$^*$} &
  &
   \multicolumn{1}{c}{\textbf{85.0\%}} &
  \multicolumn{1}{c}{60.7\%$^*$} &
  \multicolumn{1}{c}{69.4\%$^*$} 
  \\ 

   \midrule
   
\multirow{4}{*}{{Llama2-13B}} &
  \textit{0} &
  
  \multicolumn{1}{c}{\textbf{89.9\%}} &
  \multicolumn{1}{c}{87.1\%$^*$} &
  \multicolumn{1}{c}{88.8\%$^*$} &
  &
  \multicolumn{1}{c}{\textbf{96.5\%}} &
  \multicolumn{1}{c}{95.3\%$^*$} &
  \multicolumn{1}{c}{93.3\%$^*$} 
  
  \\ 

  &
  \textit{0.25} &
 
  \multicolumn{1}{c}{\textbf{74.8\%}} &
  \multicolumn{1}{c}{70.6\%$^*$} &
  \multicolumn{1}{c}{67.6\%$^*$} &
  &
  \multicolumn{1}{c}{\textbf{78.5\%}} &
  \multicolumn{1}{c}{73.9\%$^*$} &
  \multicolumn{1}{c}{67.7\%$^*$} 
  \\

  &
  \textit{0.5} &

  \multicolumn{1}{c}{\textbf{63.5\%}} &
  \multicolumn{1}{c}{59.2\%$^*$} &
  \multicolumn{1}{c}{56.5\%$^*$} &
  &
   \multicolumn{1}{c}{\textbf{57.9\%}} &
  \multicolumn{1}{c}{52.0\%$^*$}  &
  \multicolumn{1}{c}{52.7\%$^*$} \\

  &
  \textit{0.75} &

  \multicolumn{1}{c}{\textbf{57.0\%}} &
  \multicolumn{1}{c}{42.1\%$^*$} &
  \multicolumn{1}{c}{44.9\%$^*$} &
  &
   \multicolumn{1}{c}{\textbf{49.7\%}} &
  \multicolumn{1}{c}{34.9\%$^*$} &
  \multicolumn{1}{c}{41.8\%$^*$} 
  \\ 

   \midrule
\multirow{4}{*}{{Llama3-70B}} &
\textit{0} &
  
  \multicolumn{1}{c}{\textbf{92.5\%}} &
  \multicolumn{1}{c}{76.8\%$^*$} &
  \multicolumn{1}{c}{74.5\%$^*$} &
  &
  \multicolumn{1}{c}{\textbf{95.7\%}} &
  \multicolumn{1}{c}{79.5\%$^*$} &
  \multicolumn{1}{c}{73.3\%$^*$} 
  
  \\ 

  &
  \textit{0.25} &
 
  \multicolumn{1}{c}{\textbf{87.4\%}} &
  \multicolumn{1}{c}{71.3\%$^*$} &
  \multicolumn{1}{c}{67.3\%$^*$} &
  &
  \multicolumn{1}{c}{\textbf{93.1\%}} &
  \multicolumn{1}{c}{72.6\%$^*$} &
  \multicolumn{1}{c}{61.2\%$^*$} 
  \\

  &
  \textit{0.5} &

  \multicolumn{1}{c}{\textbf{82.1\%}} &
  \multicolumn{1}{c}{64.8\%$^*$} &
  \multicolumn{1}{c}{62.5\%$^*$} &
  &
   \multicolumn{1}{c}{\textbf{88.3\%}} &
  \multicolumn{1}{c}{64.1\%$^*$} &
  \multicolumn{1}{c}{55.8\%$^*$} \\

  &
  \textit{0.75} &

  \multicolumn{1}{c}{\textbf{84.0\%}} &
  \multicolumn{1}{c}{59.7\%$^*$} &
  \multicolumn{1}{c}{57.6\%$^*$} &
  &
   \multicolumn{1}{c}{\textbf{85.6\%}} &
  \multicolumn{1}{c}{56.5\%$^*$} &
  \multicolumn{1}{c}{52.4\%$^*$} 
  
   \\
   
    \midrule
\multirow{4}{*}{{Qwen2.5-32B}} &
\textit{0} &
  
  \multicolumn{1}{c}{\textbf{87.8\%}} &
  \multicolumn{1}{c}{71.3\%$^*$} &
  \multicolumn{1}{c}{75.7\%$^*$} &
  &
  \multicolumn{1}{c}{\textbf{90.7\%}} &
  \multicolumn{1}{c}{53.1\%$^*$} &
  \multicolumn{1}{c}{67.0\%$^*$}
  
  \\ 

  &
  \textit{0.25} &
 
  \multicolumn{1}{c}{\textbf{95.1\%}} &
  \multicolumn{1}{c}{79.5\%$^*$} &
   \multicolumn{1}{c}{83.4\%$^*$} &
  &
  \multicolumn{1}{c}{\textbf{97.4\%}} &
  \multicolumn{1}{c}{63.5\%$^*$} &
   \multicolumn{1}{c}{75.4\%$^*$}
  \\

  &
  \textit{0.5} &

  \multicolumn{1}{c}{\textbf{88.5\%}} &
  \multicolumn{1}{c}{72.3\%$^*$} &
    \multicolumn{1}{c}{71.7\%$^*$} &
  &
   \multicolumn{1}{c}{\textbf{92.1\%}} &
  \multicolumn{1}{c}{40.6\%$^*$} &
  \multicolumn{1}{c}{64.5\%$^*$}
  \\

  &
  \textit{0.75} &

  \multicolumn{1}{c}{\textbf{83.0\%}} &
  \multicolumn{1}{c}{66.0\%$^*$} &
   \multicolumn{1}{c}{67.3\%$^*$} &
  &
   \multicolumn{1}{c}{\textbf{86.9\%}} &
  \multicolumn{1}{c}{39.6\%$^*$} &
  \multicolumn{1}{c}{55.0\%$^*$}
  \\
   \bottomrule
\end{tabular}
\begin{tablenotes}
\item[*] indicates statistical significance compared to CoE (p < 0.05)
\end{tablenotes}
        \end{threeparttable}
}
\end{table}
Table \ref{tab:RQ3_robust} shows ACC after adding inaccurate information and produce knowledge conflicts with CoE and two types of Non-CoE.
The results show that the average ACC reaches 84.1\% for the CoE group, which is 21.4\% and 15.3\% higher than $\text{Non-CoE}_{SenP}$ and $\text{Non-CoE}_{WordP}$, respectively.
These results demonstrate CoE's superior ability in maintaining correct output when faced with conflicting information.
Furthermore, as the ratio increases from 0\% to 75\%, CoE's ACC decreases by 18.0\%, while Non-CoE variants show greater drops (24.2\% for $\text{Non-CoE}_{SenP}$ and 24.3\% for $\text{Non-CoE}_{WordP}$). 
This indicates CoE's more stable performance against conflicting knowledge. 
Considering certain context poisoning methods (such as PoisonRAG \cite{zou2024poisonedrag}) which involve injecting multiple pieces of incorrect knowledge into the context, resulting in a high conflict ratio, CoE can also better help LLMs resist such attacks.


Ablation studies demonstrate the crucial role of evidence relations in handling conflicting knowledge, with their removal leading to a substantial 62.1\% ACC drop when contradictory information is present. By connecting nodes and maintaining logical consistency, evidence relations make LLMs more resilient to contradictions. Detailed analysis is provided in Appendix \ref{sec:appendix_ablation}.


\textbf{Summary:}
When inaccurate information exists in the context, CoE can help LLMs effectively maintain the robustness against such interference. 
This suggests that existing RAG defense methods could benefit from incorporating such structured evidence chains to enhance their robustness against misleading information.

\section{Usability Assessment}
\label{sec:application}



Based on the above findings, we selected three representative tasks that leverages external knowledge, i.e., ‌RAG-based multi-hop QA, external knowledge poisoning, and poisoning defenses \cite{zhou2024trustworthiness}. 
For each task, we chodse the corresponding SOTA or prevalent baseline and modified certain components under the guidance of CoE-oriented findings to explore whether CoE-enhanced variants could achieve performance improvements.




\subsection{RAG-based Multi-Hop QA}

Given the complexity of multi-hop QA, RAG has emerged as a prevalent way for addressing such problems.
A prevalent RAG framework follows a retrieve-rank-generate pipeline: first retrieving relevant knowledge snippets using a search engine, then employing a reranker model\footnote{https://huggingface.co/BAAI/bge-reranker-large} to rank snippets based on relevance to the question, and finally using the ranked snippets as context for LLM generation. 
We select this standard RAG approach as our baseline because it represents the mainstream implementation of current RAG systems \cite{chen2024benchmarking}, which retrieve top-5 snippets from the Google Search API as context for the generation of LLM answers.

Based on this prevalent RAG framework, our variant primarily enhances the reranking process to introduce more CoE-compliant external knowledge into the context.
Specifically, while the original reranker focuses on pure relevance matching, CoE features from questions can provide additional structural guidance, as shown in Figure \ref{fig:loopjudgenew}.
\begin{itemize}
\item \textbf{CoE Feature Judgment:} The CoE features (intent, evidence nodes and relations) extracted from questions can be used to judge their presence in knowledge snippets through feature discrimination, producing judgments on feature coverage.
\item \textbf{Coverage-based Selection:} The reranking process can prioritize snippets containing more CoE features, particularly focusing on intent coverage first, followed by evidence relations and nodes. The detailed selection process is shown in Appendix \ref{sec:Algorithm}.
\end{itemize}
This optimized snippet selection serves as enhanced context for LLM generation.

\subsection{External Knowledge Poisoning}

External knowledge poisoning attacks aim to manipulate RAG systems by injecting malicious content into the knowledge base. 
We select PoisonedRAG (PR) \cite{zou2024poisonedrag} as our baseline, which uses LLM to generate false supporting documents for incorrect answers and injects them into the RAG knowledge base, causing the retriever to select these poisoned documents as context and mislead LLM to generate target answers.

Building upon PR, the SOTA knowledge poisoning attack, our variant enhances the document generation process by incorporating CoE features from target questions. By extracting CoE features from questions and integrating them into the generation process, the generated false knowledge exhibits stronger logical and semantic alignment with the questions.
The detailed generation prompts incorporating these structural features are provided in Appendix \ref{sec:appendix_coegenerate}.
\begin{table}
\centering
\resizebox{0.5\textwidth}{!}{
\begin{tabular}{c|cc|cc|cc}
\hline
\multirow{2}{*}{\textbf{Model}} & \multicolumn{2}{c}{\textbf{Multi-Hop QA(ACC)}} & \multicolumn{2}{|c|}{\textbf{Attack (ASR)}} & \multicolumn{2}{c}{\textbf{Defense (ACC/ASR)}} \\
\cline{2-7}
& RAG & RAG+CoE & PR & PR+CoE & IR & IR+CoE \\
\hline
GPT-3.5 & 68.1\% & 76.0\% & 69.0\% &79.0\% & 49.0\%/42.0\% & 78.0\%/8.0\% \\
GPT-4 & 72.9\% & 82.6\% & 49.0\% & 62.0\% & 56.0\%/38.0\% & 80.0\%/5.0\% \\
Llama2-13B & 64.4\% & 74.1\% & 62.0\% & 71.0\% & 45.0\%/51.0\% & 79.0\%/6.0\% \\
Llama3-70B & 67.8\% & 79.5\% & 60.0\% & 76.0\% & 60.0\%/37.0\% & 84.0\%/6.0\% \\
Qwen2.5-32B & 63.8\% & 77.0\% & 73.0\% & 80.0\% & 51.0\%/42.0\% & 76.0\%/6.0\% \\
\hline
\end{tabular}
}
\caption{Performance comparison between baselines and after adding CoE in three application scenarios.}
\label{tab:results}
\end{table}
\subsection{Poisoning Defense}

Poisoning defenses aim to protect RAG systems against knowledge poisoning attacks. We select InstructRAG (IR) \cite{wei2024instructrag} as our baseline, which asks LLMs to first rationalize evidence relevance and uses these rationales for context selection, enhancing the system's ability to identify and reject misleading information.


Building upon IR, the SOTA RAG defense framework, our variant strengthens its defensive capability by incorporating CoE-structured knowledge validation. By generating knowledge containing CoE features extracted from questions, these CoE-structured knowledge are injected into the knowledge base. It provides more systematic supporting evidence when the framework requests document rationales. This enables LLMs to effectively select defensive knowledge pieces within the framework. The detailed generation process is provided in Appendix \ref{sec:appendix_coegenerate}.




\subsection{Evaluation and Results}

We evaluate the effectiveness of CoE across three RAG scenarios using the HotpotQA dataset. 
To investigate how CoE enhances RAG performance in multi-hop QA, we measure the effectiveness using accuracy (ACC).
Table \ref{tab:results} shows that RAG+CoE achieves an average ACC improvement of 10.4\% compared to RAG.
Notably, RAG+CoE with GPT-4 achieves the highest ACC of 82.6\%, while Qwen2.5-32B shows the most significant improvement when CoE is integrated into RAG, with a 13.2\% increase in ACC.
The results illustrate that knowledge structured through CoE provides more effective context for LLMs to reason and generate accurate responses.

In the knowledge poisoning attack,  we measure the attack effectiveness using attack success rate (ASR).
PR+CoE achieves 11.0\% higher ASR on average across LLMs compared to PR, revealing that malicious knowledge deliberately structured through CoE becomes more effective at manipulating LLM outputs.

For RAG defense evaluation, we examine both ACC and ASR. IR+CoE demonstrates strong defensive capability with 27.2\% higher ACC and 35.8\% lower ASR compared to IR, indicating that CoE-structured defensive knowledge enables LLMs to better identify and resist misleading information while maintaining accurate responses.

\section{Conclusion}
\label{sec:Conclusion}
In this paper, we introduce CoE and investigate its impact on LLMs in imperfect external knowledge for multi-hop QA.
We characterize the CoE features and propose a discrimination approach to judge whether external knowledge exhibits the features within the user question.
Generally, our study reveals that external knowledge aligned with CoE features exhibits stronger significance, deceptiveness, and robustness against extraneous and inaccurate information in contexts.
We further validate the CoE-oriented findings by applying them to tasks that leverage external knowledge, demonstrating that the CoE-enhanced variants consistently outperform their original baseline counterparts.

\section*{Limitations}
\label{sec:Limitation}
There are three limitations to the current study.
Firstly, we apply the {\tool} to search for CoE in external knowledge, but there is no step to verify the correctness of answers within the CoE. 
If the retrieved CoE contains incorrect information, it may mislead the LLM to generate inaccurate responses.
In Section \ref{sec:faithfulness}, we discuss LLMs' Following Rate to CoE containing factual errors, showing that LLMs are highly likely to follow the knowledge provided in CoE.


Secondly, the usability of our proposed CoE-based reranking strategy ({\tool}) has inherent constraints across RAG scenarios.
For instance, some RAG scenarios convert external knowledge into vectors and store them in vector databases, then search for question-relevant knowledge at the vector level during the retrieval phase..
Our approach, which operates at the textual level, is not suitable for such vector-based RAG scenarios.

Thirdly, our approach relies on prompt-based extraction of evidence nodes using GPT-4o, potential extraction errors (either incorrect identification or missing of evidence nodes) may affect CoE's performance. We systematically analyze these scenarios in Appendix \ref{sec:appendix_keywordacc}, where experiments on 1,000 HotpotQA samples demonstrate the robustness of our method: even under imperfect extraction conditions, the accuracy only drops marginally (from 90.2\% to 89.3\% and 89.4\%). This suggests that while evidence node extraction quality matters, our approach maintains strong performance even with occasional extraction imperfections.
\section*{Ethical Statement}
Our exploration of CoE-enhanced knowledge poisoning attacks is conducted strictly for red-team testing purposes to identify and understand potential vulnerabilities in RAG systems. Following responsible security research practices, we have promptly reported our findings to relevant RAG system providers and knowledge base platforms. We present only high-level methodological insights necessary for academic understanding, without releasing detailed attack implementations. Our goal is to help develop more robust RAG systems by revealing potential weaknesses in their knowledge base integration, thereby contributing to improved security measures rather than facilitating malicious exploits.
\bibliographystyle{acl_natbib}
\bibliography{ref}
\appendix

\section{The details of the subject dataset with CoE and
two types of Non-CoE}
he details of the subject dataset with CoE and
two types of Non-CoE is shown in Table \ref{tab:pre_num}.
\begin{table}[h] \small
  \caption{The details of the subject dataset with CoE and two types of Non-CoE.}
  \label{tab:pre_num}
  
\resizebox{0.5\textwidth }{!}{

\begin{tabular}{cc|cc}
\toprule

\multicolumn{1}{c}{\textbf{Dataset}} &
\multicolumn{1}{c|}{\textbf{Type}} &
  \multicolumn{1}{c}{\textbf{Sample Num}} & 
  \multicolumn{1}{c}{\textbf{Knowledge Piece Num}}
  \\   
  
  \midrule
  \multirow{3}{*}{\textbf{HotpotQA}} &
\multirow{1}{*}{{CoE}} &
  \multicolumn{1}{c}{676} &
  \multicolumn{1}{c}{4.0} 
  \\ 
 &
\multirow{1}{*}{{SenP}} &
  \multicolumn{1}{c}{676} &
  \multicolumn{1}{c}{2.1} 
  \\ 
   &
\multirow{1}{*}{{WordP}} &
  \multicolumn{1}{c}{676} &
  \multicolumn{1}{c}{4.0} 
  \\ 
\midrule
\multirow{3}{*}{{\textbf{2WikiMultihopQA} }} &
\multirow{1}{*}{{CoE }} &
  \multicolumn{1}{c}{660} &
  \multicolumn{1}{c}{3.4}  \\ 
 &
\multirow{1}{*}{{SenP}} &
  \multicolumn{1}{c}{660} &
  \multicolumn{1}{c}{1.9} 
  \\ 
   &
\multirow{1}{*}{{WordP}} &
  \multicolumn{1}{c}{660} &
  \multicolumn{1}{c}{3.4} 
  \\ 
\bottomrule
\end{tabular}

}
\end{table}

\section{Feature Effectiveness Analysis on LLM Performance}
\label{sec:appendix_ablation}

In this analysis, we examine three types of feature perturbations: WordP, Evidence RelationP (ERP), and IntentP using GPT-3.5 as our testing model on the HotpotQA dataset. WordP involves perturbing evidence node as detailed in Section \ref{sec:non-coe_construct}. ERP removes evidence relations from the external knowledge (CoE) by prompting the LLM to modify the text while preserving other features. Similarly, IntentP removes intent information from CoE while maintaining other features.
The experimental results are presented in Table \ref{tab:ablation}.

The significance analysis (RQ1) reveals that evidence relation perturbation has the least impact on LLM accuracy, followed by evidence node perturbation, and then intent perturbation. This suggests that intent information plays the most crucial role in maintaining LLM accuracy.

Regarding deceptiveness (RQ2), CoE achieves the highest attack success rate when lacking evidence nodes, followed by missing evidence relation, and then intent.
This highlights the significance of relationships and intent in affecting LLM vulnerability.
However, CoE lacking evidence nodes demonstrates weaker stability against irrelevant external knowledge under attack scenarios, indicating that evidence nodes play a vital role in maintaining attack effectiveness when facing noisy knowledge during deception attempts.

For robustness against misinformation (RQ3), the absence of evidence relations leads to the most significant decrease in LLM accuracy when misleading information is introduced. This underscores that evidence relations are crucial features for constructing complete evidence chains and maintaining model reliability.
In conclusion, each feature demonstrates distinct strengths in different scenarios: intent information is crucial for maintaining overall accuracy, relationships are vital for constructing evidence chains and misinformation resistance, while evidence nodes play a key role in handling irrelevant knowledge under misinformation scenarios. This diverse functionality suggests that intent, evidence relations and evidence nodes are all indispensable components in constructing effective Chain-of-Evidence (CoE) for robust LLM performance.

\begin{table}[t]
\centering
\caption{Performance of GPT-3.5 with CoE and Non-CoE on HotpotQA Dataset}
\label{tab:ablation}
\resizebox{0.5\textwidth }{!}{
\begin{tabular}{c|cc|ccccc}
\toprule
RQ & Metric & Proportion Type & Proportion & CoE & WordP & ERP & IntentP \\
\toprule
\multirow{4}{*}{RQ1} & \multirow{4}{*}{ACC} & \multirow{4}{*}{Irrelevant} & 0 & 91.9\% & 79.1\% & 81.1\% & 59.9\%\\
& & & 0.25 & 90.3\% & 77.5\% & 81.6\% & 56.5\%\\
& & & 0.50 & 89.9\% & 75.4\% & 78.5\% & 54.5\%\\
& & & 0.75 & 88.9\% & 74.5\% & 76.9\% & 54.5\%\\
\midrule
\multirow{4}{*}{RQ2} & \multirow{4}{*}{ASR} & \multirow{4}{*}{Irrelevant} & 0 & 86.1\% & 83.1\% & 69.2\% & 57.3\%\\
& & & 0.25 & 85.8\% & 79.1\% & 64.8\% & 54.8\%\\
& & & 0.50 & 84.7\% & 77.8\% & 61.4\% & 53.2\%\\
& & & 0.75 & 78.4\% & 73.7\% & 58.1\% & 49.0\%\\
\midrule
\multirow{4}{*}{RQ3} & \multirow{4}{*}{ACC} & \multirow{4}{*}{Misinformation} & 0 & 91.9\% & 79.1\% & 81.1\% & 59.9\%\\
& & & 0.25 & 81.8\% & 64.0\% & 21.7\% & 53.1\%\\
& & & 0.50 & 82.0\% & 65.7\% & 21.2\% & 52.1\%\\
& & & 0.75 & 75.7\% & 60.8\% & 19.0\% & 47.8\%\\
\bottomrule
\end{tabular}
}
\end{table}

\section{ Effectiveness of CoE in Single-hop QA and Analysis of Hop Numbers}
\label{sec:single-hop}

To provide a comprehensive evaluation of CoE's effectiveness, we conducted additional experiments on single-hop scenarios alongside our main multi-hop experiments. Multi-hop questions are particularly challenging for LLMs as they require sophisticated knowledge integration and logical reasoning capabilities. However, examining single-hop scenarios helps establish the generalizability of our approach across different reasoning complexity levels.

We evaluated GPT-3.5 on a single-hop dataset (RGB) following the experimental settings from RQ1-RQ3. The results shown in Table \ref{tab:single_hop} reveal several interesting findings:
\begin{itemize}
    \item CoE demonstrates consistent effectiveness in both single-hop and multi-hop scenarios, as shown in RQ1. However, both CoE and Non-CoE exhibit stronger resistance to irrelevant information in single-hop scenarios, which can be attributed to the reduced complexity of single-step reasoning tasks.
    \item The core advantages of CoE observed in RQ2 and RQ3 remain consistent across both single-hop and multi-hop contexts, supporting the broader applicability of our approach.
    \item Our comparative analysis reveals that while the number of reasoning hops does not significantly impact CoE's significance and robustness, it notably affects Non-CoE. As the number of hops increases, SenP and WordP show decreased resistance to imperfect knowledge. This pattern emerges because multi-hop reasoning requires both individual knowledge comprehension and cross-hop integration, making the LLM more vulnerable to irrelevant or misleading information.
\end{itemize}
These findings further validate CoE's capability to effectively guide LLM reasoning regardless of the reasoning complexity, while highlighting its particular advantages in more challenging multi-hop scenarios.

Besides, to analyze the robustness of CoE across different reasoning complexity levels, We conduct statistical analysis based on results from Table \ref{tab:RQ1_performance} and Table \ref{tab:single_hop} on GPT-3.5's performance on questions requiring one-hop, two-hop, and three-hop reasoning while gradually introducing irrelevant knowledge. The results reveal interesting patterns across reasoning depths. For one-hop questions, CoE maintains consistently high accuracy (above 92.0\%) even with increasing irrelevant knowledge, demonstrating strong robustness in simple reasoning scenarios where direct evidence-to-answer mapping is sufficient.
The performance on two-hop questions shows more sensitivity to irrelevant knowledge, with accuracy declining from 91.0\% to 88.0\%. This suggests that intermediate reasoning steps are more vulnerable to distraction from irrelevant information. Interestingly, for three-hop questions, despite the higher reasoning complexity, the model shows better resilience than two-hop cases, maintaining accuracy above 90\% in most scenarios. This counter-intuitive improvement may be attributed to the LLM's enhanced focus when processing more complex reasoning chains.

\begin{table}[t]
\centering
\caption{Performance of GPT-3.5 with CoE and Non-CoE on Single-hop Dataset}
\label{tab:single_hop}
\resizebox{0.5\textwidth }{!}{
\begin{tabular}{c|cc|cccc}
\toprule
RQ & Metric & Proportion Type & Proportion & CoE & SenP & WordP \\
\toprule
\multirow{4}{*}{RQ1} & \multirow{4}{*}{ACC} & \multirow{4}{*}{Irrelevant} & 0 & 93.0\% & 74.0\% & 84.1\% \\
& & & 0.25 & 93.4\% & 77.9\% & 84.4\% \\
& & & 0.50 & 93.4\% & 80.2\% & 84.8\% \\
& & & 0.75 & 92.6\% & 79.8\% & 85.6\% \\
\midrule
\multirow{4}{*}{RQ2} & \multirow{4}{*}{ASR} & \multirow{4}{*}{Irrelevant} & 0 & 87.9\% & 55.0\% & 85.1\% \\
& & & 0.25 & 79.4\% & 47.4\% & 66.3\% \\
& & & 0.50 & 67.4\% & 40.4\% & 52.0\% \\
& & & 0.75 & 62.7\% & 32.7\% & 47.0\% \\
\midrule
\multirow{4}{*}{RQ3} & \multirow{4}{*}{ACC} & \multirow{4}{*}{Misinformation} & 0 & 93.0\% & 74.0\% & 84.1\% \\
& & & 0.25 & 86.8\% & 65.1\% & 74.0\% \\
& & & 0.50 & 83.3\% & 65.8\% & 67.4\% \\
& & & 0.75 & 77.5\% & 60.0\% & 60.0\% \\
\bottomrule
\end{tabular}
}
\end{table}
\begin{table}[t]
\centering
\caption{Accuracy of GPT-3.5 under Different Hop Num}
\label{tab:compare}
\resizebox{0.5\textwidth }{!}{
\begin{tabular}{c|ccc}
\toprule
Irrelevant Proportion & One-hop & Two-hop & Three-hop \\
\midrule
0 & 93.0\% & 91.0\% & 94.0\% \\
0.25 & 93.4\% & 89.0\% & 90.0\% \\
0.50 & 93.4\% & 88.0\% & 92.0\% \\
0.75 & 92.6\% & 88.0\% & 92.0\% \\

\bottomrule
\end{tabular}
}
\end{table}

\section{Reliability of automated evidence nodes extraction for CoE and its impact on performance}
\label{sec:appendix_keywordacc}

In our approach, we define evidence nodes and provide few-shot examples in the prompt for GPT-4o to perform evidence node extraction. Given that automated evidence node extraction may contain errors in real-world applications, we conducted a systematic analysis of potential evidence node extraction errors. These errors primarily manifest in two ways:
1) \textbf{Extraction Errors:} incorrectly identifying intent-related content as evidence nodes;
2) \textbf{Missing Errors:} failing to extract essential evidence nodes.
For example, as shown in Figure \ref{fig:CoE_explain}, Extraction Errors would occur when "state" from the intent/question is incorrectly included in the evidence nodes, while Missing Errors would happen when essential evidence node like "CEO" are not extracted, both of which could affect the accuracy of CoE identification.
To assess the impact of these potential errors, we designed corresponding perturbation operations and simulated both error types on our test dataset. The detailed experimental results and analysis are presented in Table \ref{tab:compare}.

To examine the impact of imperfect extraction, we conducted experiments on 1,000 HotpotQA samples by either adding a shared entity from intent/question (Extraction Errors) or randomly removing one evidence node (Missing Errors). 
The result show that Missing Errors led to over-identification of CoE (803 vs. 676 Our), while Extraction Errors resulted in under-identification (641).
Both scenarios slightly decreased response accuracy compared to normal conditions (90.2\% Our, 89.4\% Missing Errors, 89.3\% Extraction Errors).

\begin{table}[t]
\centering
\caption{Accuracy of GPT-3.5 under Different Evidence Nodes Error Types}
\label{tab:compare}
\resizebox{0.5\textwidth }{!}{
\begin{tabular}{c|ccc}
\toprule
Irrelevant Proportion & Our & Missing Errors & Extraction Errors \\
\midrule
0 & 91.9\% & 91.2\% & 91.2\% \\
0.25 & 90.3\% & 90.1\% & 90.1\% \\
0.50 & 89.9\% & 89.2\% & 88.4\% \\
0.75 & 88.9\% & 87.4\% & 87.5\% \\
\midrule
Num & 676 & 803 & 641 \\
\bottomrule
\end{tabular}
}
\end{table}

\section{Perturbation Strategies for Non-CoE Construction}
\label{sec:appendix_perturbation}
\begin{figure}[t!]
\centering
\setlength{\abovecaptionskip}{5pt}   
  \setlength{\belowcaptionskip}{0pt} 
\includegraphics[width=6.9cm,height=4.6cm]{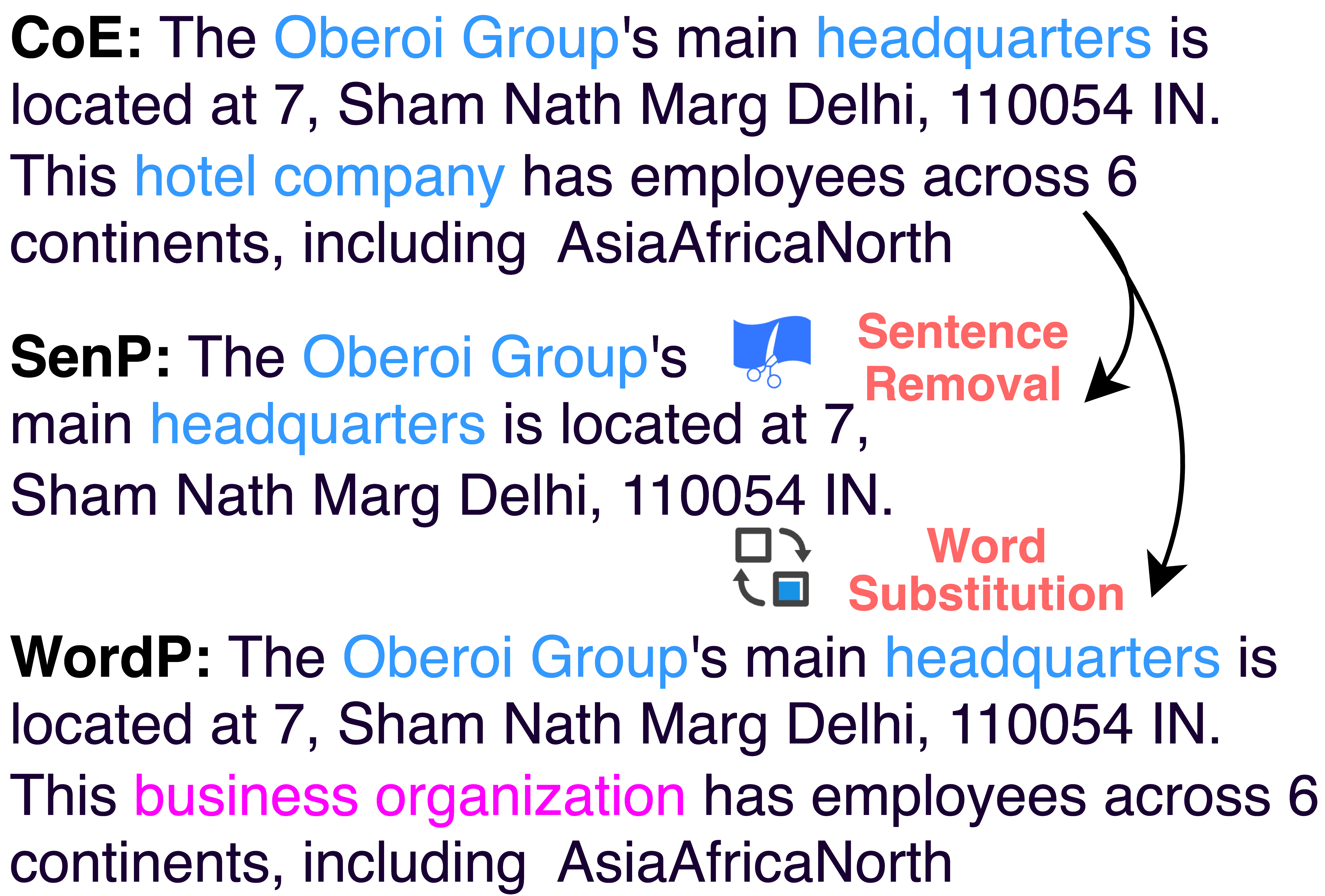}
\caption{
Examples of CoE and two types of Non-CoE.
}
\label{fig:incomplete_gen}
\end{figure}

We employ two controlled perturbation strategies to construct Non-CoE samples while maintaining the same question context:

\textbf{Sentence-Level Perturbation (SenP).}
For multihop QA, we simulate incomplete knowledge scenarios by removing knowledge pieces from CoE. We segment CoE into sentences and identify candidates containing question-mentioned evidence nodes (excluding answer nodes). We iteratively remove these candidates until CoE discrimination confirms that the remaining knowledge no longer contains complete CoE. This sentence-level approach helps understand how LLMs behave when key evidence pieces are entirely missing within the same reasoning context. Figure \ref{fig:incomplete_gen} shows this sentence-level perturbation process.

\textbf{Word-Level Perturbation (WordP).}
We create Non-CoE by replacing specific evidence nodes with their GPT-4 generated higher-level expressions (e.g., replacing hotel company'' with business organization''), maintaining more original information compared to sentence removal.
This finer-grained approach examines LLMs' sensitivity to evidence nodes while preserving most of the original semantic information. Figure \ref{fig:incomplete_gen} demonstrates this word-level perturbation approach.

\section{The  Algorithm for the Coverage-Based Selection}
\label{sec:Algorithm}
We show the detailed algorithm \ref{alg:minimal_coverage} for the minimal coverage search in {\tool}.

\begin{algorithm}[t]
\caption{Coverage-Based Selection}
\footnotesize
\label{alg:minimal_coverage}
\KwIn{External knowledge list $EK$, 
Judged external knowledge list $IEK$, where each item contains Intent, Evidence Relations, and Evidence nodes judgments}
\KwOut{
Set $S$ of minimal coverage external knowledge}
$S \leftarrow \emptyset$\;
\# Phase 1: Intent Coverage\;
\For{$i \leftarrow 0$ \KwTo $|IEK|-1$}{
    \If{$IE[i].Intent = \text{TRUE}$}{
        $S \leftarrow S \cup \{EK[i]\}$
    }
}


\# Phase 2: Evidence Relation Coverage\;
$R_{uncovered} \leftarrow$ GetUncoveredEvidencerelation($IEK, S$)\;
\For{$r \in R_{uncovered}$}{
    \For{$i \leftarrow 0$ \KwTo $|IEK|-1$}{
        \If{$IEK[i].Evidencerelation[r] = \text{TRUE}$}{
            $S \leftarrow S \cup \{EK[i]\}$\;
            \textbf{break}\;
        }
    }
}

\# Phase 3: Evidence Node Coverage\;
$K_{uncovered} \leftarrow$ GetUncoveredEvidencenodes($IEK, S$)\;
\For{$k \in K_{uncovered}$}{
    \For{$i \leftarrow 0$ \KwTo $|IEK|-1$}{
        \If{$IEK[i].Evidencenode[k] = \text{TRUE}$}{
            $S \leftarrow S \cup \{EK[i]\}$\;
            \textbf{break}\;
        }
    }
}
\Return{S}\;
\end{algorithm}

\section{The Details in {\tool}}
\label{sec:appendix_scopecoe}

\begin{figure}[h]
\centering
\setlength{\abovecaptionskip}{5pt}   
  \setlength{\belowcaptionskip}{0pt} 
\includegraphics[width=7.8cm,height=4.0cm]{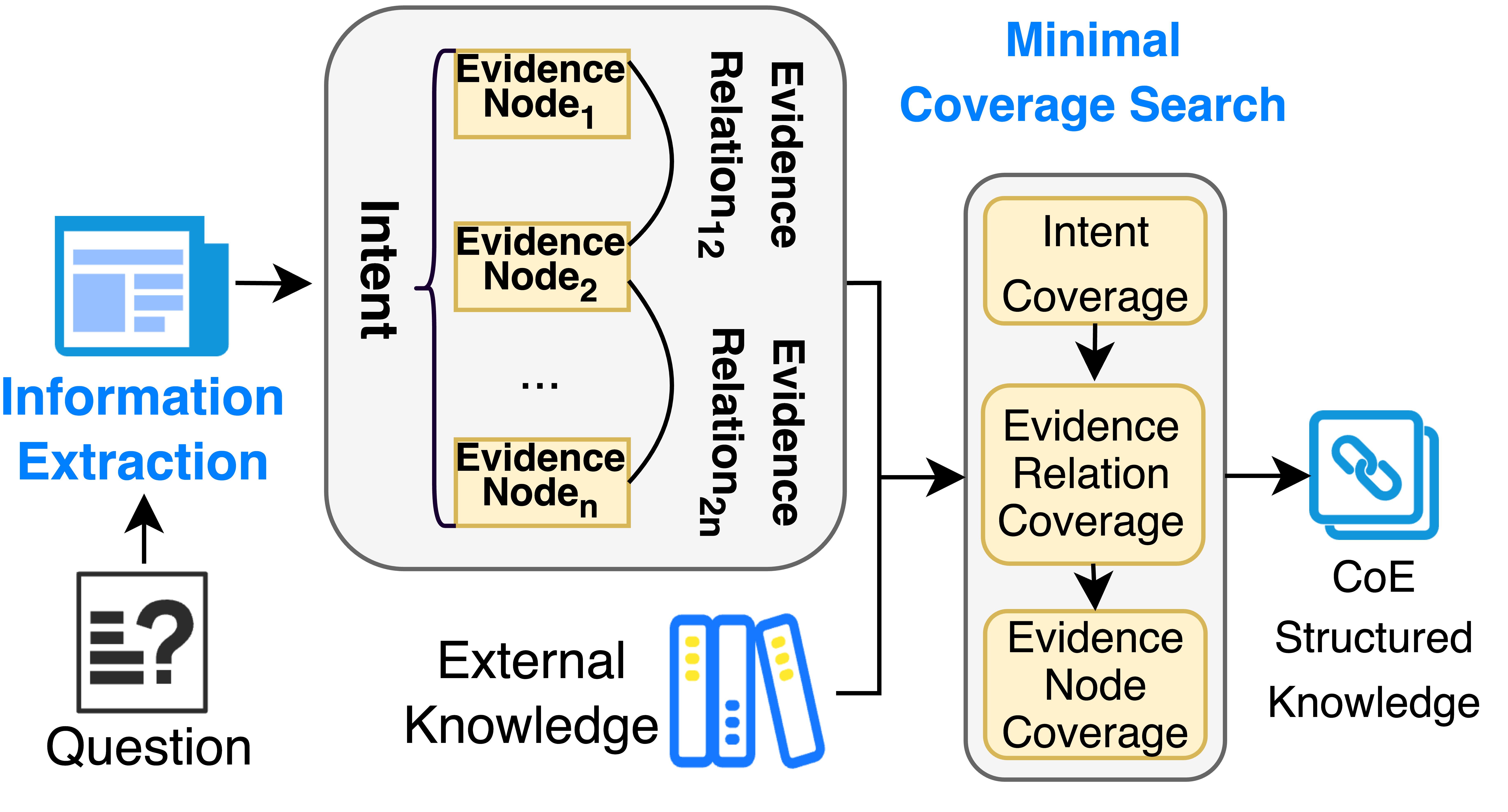}
\caption{
The overview of {\tool}. 
}
\label{fig:loopjudgenew}
\end{figure}

We show the overview of {\tool} in Figure \ref{fig:loopjudgenew}.

\label{sec:appendix_scopecoe_pipeline}

\section{Details of Information Extraction Prompts}
\label{sec:appendix_information}
The details of the information extraction prompts are illustrated below.
In pipeline, we replace the placeholders in the following prompts with the question and evidence nodes.

\begin{tcolorbox}
\small
\textbf{Intent and evidence node Extraction Prompt:}

Please extract both the intent and evidence nodes of the question, using the following criteria:

1)	As for intent, please indicate the content intent of the evidence that the question expects, without going into specific details.

2)	 As for evidence nodes, Please extract the specific details of the question.

The output must be in json format, consistent with the sample.
Here are some examples:

\textbf{Example1:}

Question:750 7th Avenue and 101 Park Avenue, are located in which city?

Output: \{ "Intent": "City address Information", "evidence nodes": ["750 7th Avenue", "101 Park Avenue"] \}

\textbf{Example2:}

Question: The Oberoi family is part of a hotel company that has a head office in what city?

Output: \{ "Intent": "City address Information", "evidence nodes": ["Oberoi family", "head office"] \}

\textbf{Example3:}

Question: What nationality was James Henry Miller's wife?

Output: \{ "Intent": "Nationality of person", "evidence nodes": ["James Henry Miller", "wife"] \}

\textbf{Example4:}

Question: What is the length of the track where the 2013 Liqui Moly Bathurst 12 Hour was staged?

Output: \{ "Intent": "Length of track", "evidence nodes": ["2013 Liqui Moly Bathurst 12 Hour"] \}

\textbf{Example5:}

Question: In which American football game was Malcolm Smith named Most Valuable player?
Output: \{ "Intent": "Name of American football game", "evidence nodes": ["Malcolm Smith", "Most Valuable player"] \}

\textbf{Question:}  \textit{\textbf{[Question]}}
\textbf{Output:}

\end{tcolorbox}

\begin{tcolorbox}
\small
\textbf{Evidence Relations Extraction Prompt:}

Please extract evidence relations based on the input questions and evidence nodes, using the following criteria:

    1) Each evidence relation has two elements, the implied evidence nodes and the textual description of the evidence relations.
    
    2) The description of the evidence relations is limited to the two evidence nodes and does not involve other evidence nodes.
    
    3) If there is no evidence relation between evidence nodes, no extraction is required.
    
The output must be in json format, consistent with the examples. 
Here are some examples:

The output must be in json format, consistent with the sample.
Here are some examples:

\textbf{Example1:}

Question:750 7th Avenue and 101 Park Avenue, are located in which city?

Evidence nodes:["750 7th Avenue", "101 Park Avenue"]

Output: []

\textbf{Example2:}

Question: Lee Jun-fan played what character in \"The Green Hornet\" television series?

Evidence nodes:["Lee Jun-fan", "The Green Hornet"]

Output: [\{"Evidence nodes":["Lee Jun-fan", "The Green Hornet"], "Evidence Relations: "played character in"\}]

\textbf{Example3:}

Question: In which stadium do the teams owned by Myra Kraft's husband play?

Evidence nodes: ["teams", "Myra Kraft's husband"]

Output: [\{"Evidence nodes":["teams", "Myra Kraft's husband"], "Evidence Relations": "is owned by"\}]

\textbf{Example4:}

Question: The Colts' first ever draft pick was a halfback who won the Heisman Trophy in what year?

Evidence nodes:["Colts' first ever draft pick", "halfback", "Heisman Trophy"]

Output:[\{"Evidence nodes":["Colts' first ever draft pick", "halfback"], "Evidence Relations": "was"\}]

\textbf{Example5:}

Question: The Golden Globe Award winner for best actor from "Roseanne" starred along what actress in Gigantic?

Evidence nodes:["Golden Globe Award winner", "best actor", "Roseanne", "Gigantic"]

Output: [\{"Evidence nodes":["Golden Globe Award winner", "best actor"], "Evidence Relations": "for"\}, \{"Evidence nodes":["best actor", "Roseanne"], "Evidence Relations": "starred in "\}]

\textbf{Question:}  \textit{\textbf{[Question]}}

\textbf{Evidence nodes:} \textit{\textbf{[Evidence node]}}

\textbf{Output:}

\end{tcolorbox}

\section{Details of Feature Discrimination Prompts}
\label{sec:appendix_complete}
The details of the Feature Discrimination prompts are illustrated below.
In pipeline, we replace the placeholders in the following prompts with the external knowledge, intent, evidence node, and evidence relation.

\begin{tcolorbox}
\small
\textbf{Intent Discrimination Prompt:}

Please determine whether the input intent is covered in the input external knowledge. Please output only "yes" or "no".

\textbf{Input intent:}  \textit{\textbf{[Intent]}}

\textbf{Input external knowledge:} \textit{\textbf{[External Knowledge]}}

\end{tcolorbox}

\begin{tcolorbox}
\small
\textbf{Evidence Nodes Discrimination Prompt:}

Please determine if the input evidence node is mentioned in the input external knowledge. It doesn't necessarily need to be an exact character match; partial matches or semantic similarities are also acceptable. Please output only "yes" or "no".

\textbf{Input evidence node:}  \textit{\textbf{[Evidence node]}}

\textbf{Input external knowledge:} \textit{\textbf{[External Knowledge]}}

\end{tcolorbox}

\begin{tcolorbox}
\small
\textbf{Evidence Relations Discrimination Prompt:}

Please determine if the input external knowledge supports the logical relationship between the two given evidence nodes. If there is explicit evidence in the input knowledge that confirms the evidence node-evidence relation-evidence node triple, output "yes"; otherwise output "no". Please respond only with "yes" or "no".

\textbf{Input triple: \textit{(evidence node1, evidence relation, evidence node2)}}

\textbf{Input external knowledge:} \textit{\textbf{[External Knowledge]}}

\end{tcolorbox}






\section{Details of the CoE-Guided Knowledge Generation Prompts}
\label{sec:appendix_coegenerate}
\begin{tcolorbox}
\small
\textbf{CoE-Guided Knowledge Generation Prompt:}

Given:  

\textbf{        Evidence Nodes: [Nodes]}  
        
       \textbf{ Evidence Relations: [Relations] } 
        
       \textbf{ Intent: [Intent]  }
        
        \textbf{Question: [Question]}  
        
       \textbf{ Answer: [Answer]  }

    Create a corpus that:  
    
    1. Integrate all given [Nodes] naturally, do not replace [Nodes] with pronouns, each sentence in the corpus include as many [Nodes] as possible. 
    
    2. If there are [Relations] between [Nodes], when constructing sentences containing these [Nodes], try to express the [Relations] between them.  
    
    3. The corpus should align with the meaning of the [Intent].  
    
    4. Flows naturally as coherent text  
    
    5. Please limited the corpus to 100 words.  
    
    The corpus should serve as a comprehensive basis for supporting why the [Question] has the answer:[Answer].  
    
    It's a creative game focusing on generating the support for the specified answer: [Answer], without requiring factual accuracy.  

\end{tcolorbox}

\section{Details of the Answer Generation Prompts}
\label{sec:appendix_answer}
The details of the Answer Generation prompts are illustrated below.
In pipeline, we replace the placeholders in the following prompts with the correct answer.

\begin{tcolorbox}
\small
\textbf{Answer Generation Prompt:}

For the input phrase, please generate a phrase of similar type and format, but not the same. Just output the phrase, no explanation is needed, the expression form is consistent with the examples. Here are some examples:

\textbf{Example1:}

Input phrase: United States

Output: Canada

\textbf{Example2:}

Input phrase: alcohol

Output: Soda

\textbf{Example3:}

Input phrase: September 29, 1784

Output: April 22, 1964

\textbf{Example4:}

Input phrase: Laura Ellen Kirk

Output: Elon Musk

\textbf{Example5:}

Input phrase: 39,134

Output: 19,203

\textbf{Input phrase:}  \textit{\textbf{[Correct Answer]}}

\textbf{Output:} 

\end{tcolorbox}

\end{document}